%% file: work-1/TOMM-Manuscript.tex
\documentclass[acmsmall]{acmart}

\usepackage{amsmath,amsfonts}
\usepackage{algorithmic}
\usepackage{algorithm}
\usepackage{array}
\usepackage[caption=false,font=normalsize,labelfont=sf,textfont=sf]{subfig}
\usepackage{textcomp}
\usepackage{stfloats}
\usepackage{url}
\usepackage{verbatim}
\usepackage{graphicx}
\usepackage{float} 
\usepackage{amsmath}
\usepackage{placeins} 
\usepackage{adjustbox} 
\usepackage{booktabs} 
\usepackage{multirow}
\usepackage{bbding}

\usepackage{makecell}
\usepackage{boldline, hhline}
\usepackage{colortbl}

\definecolor{mygray}{RGB}{230,230,230}
\definecolor{graytxt}{RGB}{246,240,233}
\definecolor{bluetxt}{RGB}{53,160,219}
\definecolor{bluegray}{RGB}{241,244,250}
\definecolor{orangegray}{RGB}{251,244,242}
\definecolor{orange}{RGB}{248,231,228}
\definecolor{darkorange}{RGB}{238,196,188}
\definecolor{gree}{RGB}{226,239,213}
\definecolor{greegray}{RGB}{245,249,241}

\definecolor{bluelight}{RGB}{223,246,255}

\definecolor{greenright}{RGB}{1,146,65}
\definecolor{redwrong}{RGB}{204,35,40}

\usepackage{url}
\usepackage{hyperref}
\hypersetup{colorlinks=true,
            linkcolor=red,
            anchorcolor=blue,
            citecolor=blue}

\AtBeginDocument{%
  \providecommand\BibTeX{{%
    \normalfont B\kern-0.5em{\scshape i\kern-0.25em b}\kern-0.8em\TeX}}}

\setcopyright{acmcopyright}
\copyrightyear{2021}
\acmYear{2021}
\acmDOI{10.1145/1122445.1122456}

\acmJournal{TOMM}
\acmVolume{6}
\acmNumber{2s}

\begin{document}

\title{Beyond Seen Bounds: Class-Centric Polarization for Single-Domain Generalized Deep Metric Learning}

\author{Xin Yuan}
\affiliation{%
  \institution{School of Computer Science and Technology, Wuhan University of Science and Technology}
  \city{Wuhan}
  \country{China}}
   \email{xinyuan@wust.edu.cn}

\author{Meiqi Wan}
\affiliation{%
  \institution{School of Computer Science and Technology, Wuhan University of Science and Technology}
  \city{Wuhan}
  \country{China}}
  \email{wanmeiqi@wust.edu.cn}

\author{Wei Liu}
\affiliation{%
  \institution{School of Computer Science and Technology, Wuhan University of Science and Technology}
  \city{Wuhan}
  \country{China}}
  \email{liuwei@wust.edu.cn}

\author{Xin Xu}
\affiliation{%
  \institution{School of Computer Science and Technology, Wuhan University of Science and Technology}
  \city{Wuhan}
  \country{China}}
\email{xuxin@wust.edu.cn}
\authornote{Corresponding Author}

\author{Zheng Wang}
\affiliation{%
 \institution{National Engineering Research Center for Multimedia Software, Institute of Artificial Intelligence, School of Computer Science, Wuhan University}
 \city{Wuhan}
 \country{China}}
\email{wangzwhu@whu.edu.cn}


\renewcommand{\shortauthors}{Xin Yuan and et al.}

\begin{abstract}
  Single-domain generalized deep metric learning (SDG-DML) faces the dual challenge of \textbf{\textit{both category}} and \textbf{\textit{domain shifts}} during testing, limiting real-world applications.
  Therefore, aiming to learn better generalization ability on both unseen categories and domains is a realistic goal for the SDG-DML task. 
  To deliver the aspiration, existing SDG-DML methods employ the domain expansion-equalization strategy to expand the source data and generate out-of-distribution samples.
  However, these methods rely on proxy-based expansion, which tends to generate samples clustered near class proxies, failing to simulate the broad and distant domain shifts encountered in practice.
  To alleviate the problem, we propose \textbf{CenterPolar}, a novel SDG-DML framework that dynamically expands and constrains domain distributions to learn a generalizable DML model for wider target domain distributions. 
  Specifically, \textbf{CenterPolar} contains two collaborative class-centric polarization phases: (1) \textbf{C}lass-\textbf{C}entric \textbf{C}entrifugal \textbf{E}xpansion (\textbf{C$^3$E}) and (2) \textbf{C}lass-\textbf{C}entric \textbf{C}entripetal \textbf{C}onstraint (\textbf{C$^4$}). 
  In the first phase, \textbf{C$^3$E} drives the source domain distribution by shifting the source data away from class centroids using centrifugal expansion to generalize to more unseen domains.
  In the second phase, to consolidate domain-invariant class information for the generalization ability to unseen categories, \textbf{C$^4$} pulls all seen and unseen samples toward their class centroids while enforcing inter-class separation via centripetal constraint.
  Extensive experimental results on widely used CUB-200-2011 Ext., Cars196 Ext., DomainNet, PACS, and Office-Home datasets demonstrate the superiority and effectiveness of our \textbf{CenterPolar} over existing state-of-the-art methods.
  \textit{The code will be released after acceptance.}
\end{abstract}


\maketitle

\section{Introduction}
\label{sec:intro}
Deep Metric Learning (DML) aims at learning a distance metric model to measure the similarity/dissimilarity between samples~\cite{roth2022non,bhatnagar2025potential}, which has been applied in many applications, such as image retrieval~\cite{yuan2023osap,zhu2025interactive}, face recognition~\cite{wang2018cosface,sun2020circle}, and object re-identification~\cite{xiao2017margin,xu2022rank}. 
The goal of DML is to pull together intra-class samples while pushing away inter-class samples in the learned metric space. Previous studies in DML mainly focus on two aspects, \textit{i.e.}, pair- and proxy-based DML methods~\cite{kim2020proxy,bhatnagar2025potential}. Previous DML solely focuses on unseen categories within the seen domain, whereas generalization ability studies about the unseen domain have not been deliberated.
Regardless of achieving good performance on a seen domain, the model's performance will degrade dramatically on the unseen domain owing to the domain shift between the source and target domains. In real-world applications, the image retrieval system will unavoidably ﬁnd a query image in a new scenario. Therefore, learning a better generalization ability on unseen domains is critical for the DML method.

To meet this goal, out-of-distribution generalization technologies as a straightforward solution, including Domain Adaptation (DA)~\cite{zhao2025multi} and Domain Generalization (DG)~\cite{chen2023meta,zhou2024mixstyle} methods, have been proposed to improve the generalizability of the learned metric model. 
However, with training data of seen categories and domains, neither DA nor DG methods can be directly applied to generalize a DML model to testing data with different unseen categories and domains. 
The main reasons are: (1) both DA and DG are designed for the classification task, which involves unseen domains within seen categories instead of unseen domains within unseen categories; 
and (2) compared to the classification task, DML primarily focuses on learning the similarity information between data rather than the label information. The most iconic DG method is Single-Domain Generalization (SDG), which shares the same setting as DML in training data, requiring only data from a single source domain for model training.
Meanwhile, existing SDG methods~\cite{wang2021learning,zhu2022crossmatch,fan2024domain,zhao2024novel} have been designed to alleviate the impact of domain shift caused by the different distribution between training and test data.
Therefore, in contrast to earlier DML, this paper investigates Single-Domain Generalized Deep Metric Learning (SDG-DML), which addresses the dilemma of integrating SDG and DML tasks.
It faces the dual challenges, \textit{i.e.}, category shifts within unseen categories as well as domain shifts within unseen domains between training and testing data in real-world scenarios.
Consequently, SDG-DML is a more challenging yet realistic task for real-world applications.

\begin{figure}[t]
  \centering
  \includegraphics[width=\linewidth]{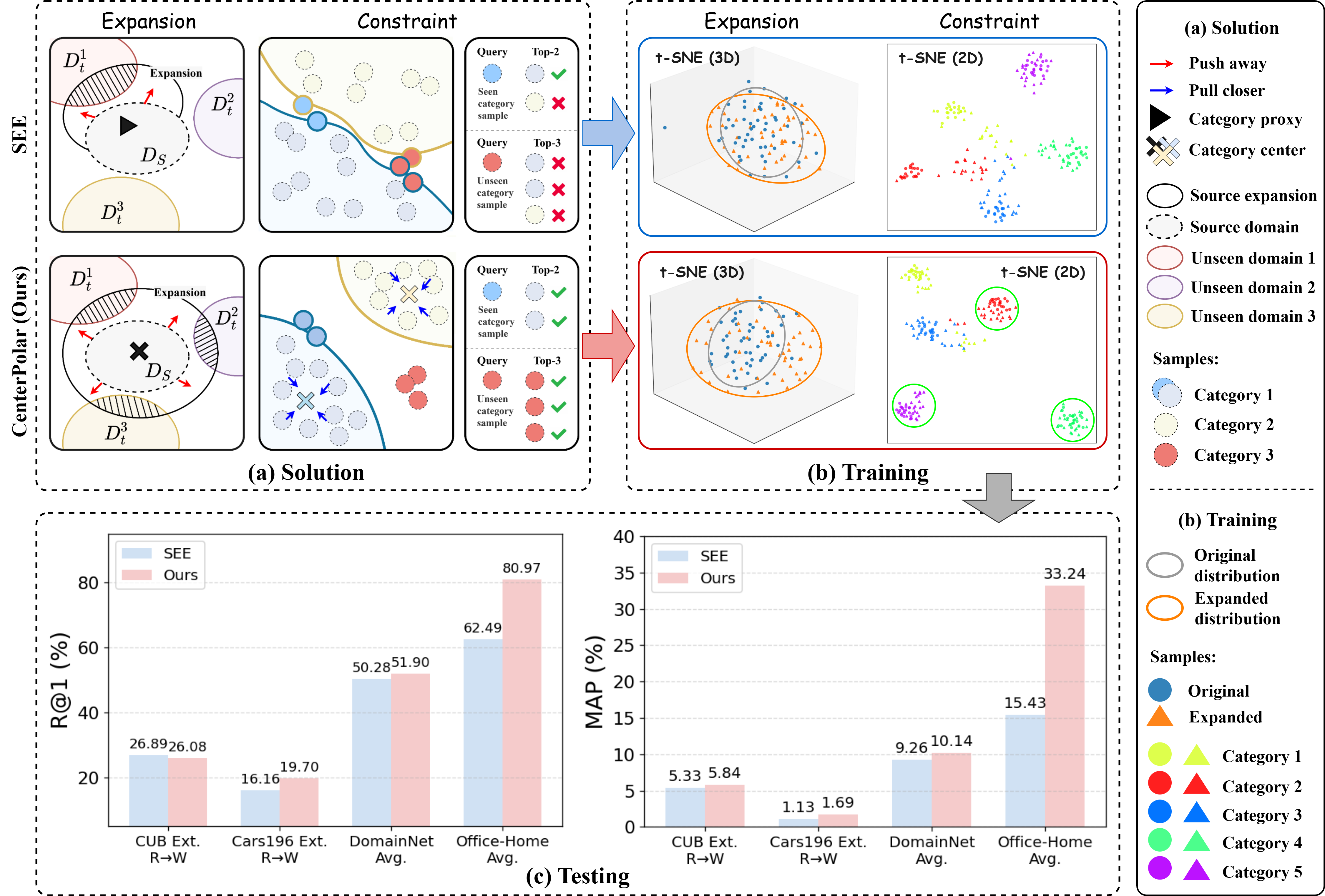}
  \caption{Comparison of different solutions in SDG-DML.
(a) Solutions: A comparison of the existing class proxy-based expansion method (\textit{i.e.}, the SEE method) with our class center-based expansion method (\textit{i.e.}, our \textbf{CenterPolar} method). (b) Training: Corresponding to the solutions, we visualized the expansion and constraint processes during training using t-SNE. (c) Testing: The test results obviously show that our method can improve the generalization of the learned metric model. (\textit{Best viewed in color}).    
  }
  \vspace{-3mm}
  \label{fig:fig1}
\end{figure}

In particular,
Yan \textit{et al.} first pioneered the SDG-DML task to advance the development of the DML field and also provided the SEE method for SDG-DML that achieves remarkable success~\cite{yan2023learning}. 
Meanwhile, up to date, 
there has been only one method in the SDG-DML task, \textit{i.e.}, SEE.
Nevertheless, as shown in Fig.~\ref{fig:fig1}(a), SEE suffers from two limitations that are extremely susceptible to collapsing the learned metric model, \textit{i,e.}, \textbf{\textit{insufficient domain expansion}} and \textbf{\textit{brittle class-discriminative constraint}}. Specifically, firstly, SEE's domain expansion is based on proxies, and the instability of these proxies~\cite{roth2022non} may lead to biased and insufficient coverage of the expanded feature space. The augmented samples are mainly shifted along the proxy positions, resulting in an unbalanced distribution that only captures a portion of the target domain. Secondly, as shown in Fig.~\ref{fig:fig1}(b), due to the lack of explicit clustering or constraints on the augmented samples, the generated features are relatively scattered in space, further weakening the model's generalization ability to distinguish between unseen categories and domains.

To overcome these challenges, this paper proposes a \textbf{CenterPolar} framework for SDG-DML, which can simulate more latent unseen target domains, as illustrated in Fig.~\ref{fig:fig1}(a).
\textbf{CenterPolar} consists of two collaborative class-centric polarization phases, \textit{i.e.}, \textbf{C}lass-\textbf{C}entric \textbf{C}entrifugal \textbf{E}xpansion (\textbf{C$^3$E}) and \textbf{C}lass-\textbf{C}entric \textbf{C}entripetal \textbf{C}onstraint (\textbf{C$^4$}). 
In \textbf{C$^3$E} phase, we controllably expand the seen source domain distribution. For each sample, we compute its class centroid and generate an augmented sample. Then, we push the augmented sample from the corresponding class centroid while leveraging a margin constraint to prevent semantic shift and expansion collapse. We hope to discover more latent unseen target domains to learn the domain-invariant metric.
Fig.~\ref{fig:fig1}(b) comparatively illustrates the training process of the two methods. On the left, it can be seen that our \textbf{CenterPolar} can effectively expand to a larger range (orange circle), while on the right, it can be observed that samples of the same class are more tightly clustered in the feature space (green circle). Fig.~\ref{fig:fig1}(c) shows the test results, demonstrating that our \textbf{CenterPolar} achieves superior performance compared to SEE.
Our contributions are as follows:

\begin{itemize}
    \item We propose a novel \textbf{CenterPolar} framework for SDG-DML, 
    which can simulate more latent unseen target domains via dynamically expanding and constraining the domain distribution. 
    Our framework enables the metric model to learn a class-discriminative and domain-invariant distance metric for unseen categories and domains, and thereby improves the generalizability of the learned metric model.

    \item We present two collaborative class-centric polarization phases: \textbf{C$^3$E} and \textbf{C$^4$}. \textbf{C$^3$E} employs centrifugal expansion, pushing source data away from class centroids to broaden domain distributions and learn domain-invariant metric for unseen domains. \textbf{C$^4$} applies centripetal constraint, pulling all samples towards their class centroids, which ensures intra-class compactness and inter-class separation. This solidifies class-discriminative metric for generalizing to unseen categories.

    \item Our experiments on five widely used SDG-DML benchmarks show that our proposed method achieves competitive performance over existing state-of-the-art DML and SDG-DML methods. 
    Notably, compared to the existing SDG-DML method, \textit{i.e.}, SEE, our method also exhibits significant advantages in 
    training time 
\end{itemize}

The rest of this paper is organized as follows: \S~\ref{sec: related work} reviews related works. Then, \S~\ref{sec:methodology} introduces our method in this paper. Next, we conduct experiments to demonstrate the superiority and effectiveness of our \textbf{CenterPolar} in \S~\ref{sec:experiments}. Finally, we make the corresponding conclusions of this work in \S~\ref{sec:conclusion}.


\section{Related Work}
\label{sec: related work}
In this section, we first review the deep metric learning (DML) methods in \S~\ref{sec: dml}. Then, \S~\ref{sec: oodg for dml} primarily concentrates on the out-of-distribution generalization for DML methods. Finally, \S~\ref{sec:da} mainly discusses data augmentation methods in the field of DML and SDG.


\subsection{Deep Metric Learning}
\label{sec: dml}
Deep Metric Learning (DML) methods are categorized into pair-based and proxy-based DML methods.
The pair-based DML methods learn the distance/similarity relationship between samples by building pairs or tuples of samples.
Early, Hadsell \textit{et al.} proposed Contrastive Loss~\cite{hadsell2006dimensionality} to learn similarity and dissimilarity pairs for preventing the system collapse. 
After that, researchers introduce an anchor to constitute a triplet or tuple to constrain positive and negative samples, \textit{i.e.}, Triplet Loss~\cite{schroff2015facenet,wu2017sampling}, Quadruplet Loss~\cite{xiao2017margin}, and their variants~\cite{wang2019multi,sun2020circle}.
Despite its impressive success, the pair-based DML methods suffer from slow convergence due to the computational cost of numerous sample pair relations.

To enable fast and reliable convergence, the proxy-based DML methods reduce the complexity by introducing proxies. 
Early, ProxyNCA~\cite{movshovitz2017no} and its variants~\cite{teh2020proxynca++,kirchhof2022non} assign a proxy for each class and associate proxies with samples, then pull positive pairs of samples closer while pushing negative pairs farther away. 
To consider the rich data-to-data relations, ProxyAnchor~\cite{kim2020proxy} combines the advantages of both pair- and proxy-based DML methods. Yet, these methods ignore complex intra-class situations, thus failing to capture discriminative relationships among data.
Some works introduce multiple proxies per class to better model both within- and inter-class relationships~\cite{qian2019softtriple,zhu2020fewer}.
However, all the above methods still suffer from performance degradation on unseen domains, due to the existence of both category and domain shifts. 
To address the above challenges, Yan \textit{et al.} first proposed the SDG-DML task, and introduced SEE to tackle this task~\cite{yan2023learning}, its generalization ability for unseen categories and unseen domains is still restricted due to insufficient domain expansion and brittle class discriminative constraint.

\subsection{Out-of-Distribution Generalization} 
\label{sec: oodg for dml}
Existing out-of-distribution generalization (OODG) methods are classified into domain adaptation (DA) and domain generalization (DG) according to whether or not access to the target domain. 
On the one hand, DA alleviates the domain shift problem by reducing the feature distribution gap between the source and target datasets~\cite{zhou2024mixstyle,zhao2025multi}. 
For the DML task, Geng \textit{et al.} proposed the DA metric learning by applying data-dependent regularization~\cite{geng2011daml}.
Sohn \textit{et al.} leveraged the unsupervised DA metric learning for the disjoint classification task via aligning them with an auxiliary domain of transformed source domain features~\cite{sohn2018unsupervised}.
Ren \textit{et al.} utilized DA to align the
sample and proxy distributions for improving the performance of proxy-based DML.
Since these methods~\cite{ren2024towards,zhao2025multi} are specifically designed for the classification task and can access target domain data, they are different from DML settings, and therefore cannot be directly applied to the DML.

On the other hand, compared to DA, DG has no access to any target domain, which is more suitable for the DML setting. 
Milbich \textit{et al.} designed a novel triplet sampling strategy that can be easily applied on top of recent ranking loss frameworks~\cite{milbich2020sharing}.
Katsumata \textit{et al.} decoupling loss that diffuses the feature representations of unknown samples~\cite{katsumata2021open}.
Wang \textit{et al.} applied the implicit semantic augmentation strategy for DML to expand class diversity by exploring holistic representations of the class distribution~\cite{wang2022semantic}. 
Recently, Single-Domain Generalization (SDG), a more practical scenario aligned with typical DML settings, has emerged to tackle domain shift using only one source domain. Existing SDG methods can be broadly categorized into two main types~\cite{yang2024causality,fan2024domain,zhao2024novel,chi2025contrast,qu2023modality,wan2022meta,wang2025progressive}. The first type focuses on domain augmentation, explicitly constructing diverse data distributions within a single source domain to simulate potential domain shifts and improve model robustness on unseen domains. For example, some works perform domain expansion from a causal perspective or expand feature coverage by designing perturbation strategies \cite{yang2024causality,fan2024domain,zhao2024novel}. The second type emphasizes feature disentanglement, no longer directly generating new samples, but instead focusing on extracting domain-invariant causal or structural information from existing data. Common approaches include contrastive learning, debiasing learning, and meta-learning strategies \cite{chi2025contrast,qu2023modality,wan2022meta}. Furthermore, some research attempts to combine both augmentation and disentanglement, simultaneously expanding domain diversity while explicitly constraining the model to focus on domain-invariant features, thereby further improving generalization capabilities \cite{wang2025progressive}.
However, these strategies fundamentally rely on explicit class labels and well‑defined decision boundaries, making them difficult to adapt directly to DML, where learning objectives are defined by pairwise or list‑wise similarities rather than classification boundaries.


Therefore, these methods still struggle with the challenge of category and domain shifts on unseen domains in SDG-DML.
Recently, Yan \textit{et al.} combined DML and SDG to form the SDG-DML task and proposed the proxy-based method, \textit{i.e.}, SEE, through adaptive domain expansion and domain-aware equalization~\cite{yan2023learning}.
In contrast to SEE, we propose to employ class-centric polarization to expand and constrain samples, thereby achieving more effective performance with lower computational cost. 

\subsection{Data Augmentation}
\label{sec:da}
Data augmentation is widely used to improve model generalization. This is particularly critical in DML and SDG tasks, where it addresses two fundamental challenges: \textbf{Challenge 
(1)} Mitigating intra-class variation in DML by enriching sample diversity to improve similarity measurement for mitigating category shift~\cite{deng2022deep,park2025deep};
\textbf{Challenge (2)} Simulating potential domain shifts in SDG by generating synthetic domains from a single-source distribution~\cite{wang2025progressive,chi2025contrast}.

Traditional augmentation techniques, including cropping, rotation, and color jittering, provide elementary variability that enhances model robustness~\cite{shorten2019survey,kim2020proxy}. However, they exhibit limited efficacy when confronting complex real-world distribution shifts. To overcome this limitation, learning-based augmentation paradigms have emerged as prominent alternatives. Feature-level augmentation focuses on diversifying latent representations rather than raw pixels, typically by reshaping feature distributions to improve robustness. Typical approaches include mixing features between samples (\textit{e.g.}, MixUp~\cite{jin2024survey}), and simulating potential domain shifts through statistical mixing or normalization perturbations (\textit{e.g.}, MixStyle~\cite{zhou2024mixstyle}, NormAug~\cite{qi2024normaug}). In addition, adversarial augmentation effectively enriches domain diversity and further improves the generalization ability of models by finding the most unfavorable perturbations in the feature space or synthesizing more challenging samples using a generator-discriminator framework~\cite{chen2020simple,cheng2023adversarial,fan2024domain}.

However, these methods frequently depend on computationally intensive frameworks, such as adversarial training or external style guidance, which introduce additional implementation complexity. Critically, insufficient constraints on sample generation may compromise training stability or prove inefficient to induce meaningful domain variability. Consequently, developing streamlined augmentation strategies becomes imperative: expected approaches that ensure semantic consistency while effectively expanding the domain coverage space through architecturally simple yet potent mechanisms. It is precisely this insight that has inspired the design of the methodology for this work.


\section{Methodology}
\label{sec:methodology}
In this section, we first introduce the preliminaries in this work in \S~\ref{sec: pre}. Next, we introduce the motivation of the proposed \textbf{CenterPolar} in \S~\ref{sub:mot}. Then, we describe our proposed \textbf{CenterPolar} framework in detail in \S~\ref{subsec:c3e} and \S~\ref{subsec:c4}. As shown in Fig.~\ref{fig:framework}, the proposed \textbf{CenterPolar} framework consists of two collaborative class-centric polarization phases, \textit{i.e.,} \textbf{C$^3$E} and \textbf{C$^4$}, which aim to effectively generalize the learned DML model to unseen categories and domains. Finally, we provide the overall procedure and detailed theoretical analysis for \textbf{CenterPolar} in \S~\ref{sec: overall} and \S~\ref{sub:theor}, respectively.

\subsection{Preliminaries}
\label{sec: pre}
\noindent \textbf{Single-Domain Generalized Deep Metric Learning.}
Single-Domain Generalized Deep Metric Learning (SDG-DML) extends Deep Metric Learning (DML) by addressing the domain shift on unseen domains. 
In SDG-DML, let $\mathcal{D}_{S} = \left\{\left(\mathbf{x}_{i}^{s}, y_{i}\right)\right\}_{i=1}^{N_s}$ and $\mathcal{D}_{T} = \left\{\left(\mathbf{x}_{j}^{t}\right)\right\}_{j=1}^{N_t}$ be a labeled source domain and an unlabeled target domain.
Here, $\left(\mathbf{x}_{i}^{s}, y_{i}\right)$ signifies $i$-th sample $\mathbf{x}_{i}^{s}$ with the corresponding category label $y_{i}^{s} \in \left\{1,2, \cdots, C\right\}$ in $\mathcal{D}_{S}$, while $\left(\mathbf{x}_{j}^{t}\right)$ denotes $j$-th sample $\mathbf{x}_{j}^{t}$ without the corresponding category label in $\mathcal{D}_{T}$. 
In particular, $\mathcal{D}_{S}$ and $\mathcal{D}_{T}$ essentially represent a training dataset with seen categories and domain, and a testing dataset with unseen categories and domain, respectively.
The goal of SDG-DML is to train a metric model $\mathcal{M}$ on $\mathcal{D}_{S}$ that generalizes well to a given unseen domain $\mathcal{D}_{T}$, \textit{i.e.}, learning a projector metric function: $f_{\mathcal{M}}: \mathcal{D}_{S} \xrightarrow{f} \mathcal{M}$, which maps the source domain $\mathcal{D}_{S}$ to a metric space $\mathcal{M}$.

Specifically, the distribution of $\mathcal{D}_{S}$ (\textit{i.e.}, $\mathcal{P}(\mathcal{D}_{S})$) differs from that of $\mathcal{D}_{T}$ (\textit{i.e.}, $\mathcal{P}(\mathcal{D}_{T})$) in SDG-DML, that is, $\mathcal{P}(\mathcal{D}_{S}) \neq \mathcal{P}(\mathcal{D}_{T})$. This distribution shift can be further detailed into the category shift and the domain shift. 
The optimization objective of SDG-DML can be defined as follows:
\begin{equation}
    \mathcal{M}_{\Theta^*} := \arg \min_{\mathcal{M}_{\Theta}} \mathbb{E}_{f_\mathcal{M}} [\frac{1}{N_s} \sum_{k=1}^{N_s} \ell (\mathcal{D}_{S}; \Theta)],
\end{equation}
where $\mathcal{M}_{\Theta^*}$ represents the final metric model and $\Theta$ indicates the network parameters. And $\mathbb{E}_{f_\mathcal{M}}$ is the empirical risk of $f_\mathcal{M}$, which aims to minimize the average loss on $\mathcal{D}_{S}$ for learning better generalization ability.




\noindent \textbf{Distance Metric.} 
Note that common distance metrics, such as Euclidean~\cite{wang2005euclidean}, Mahalanobis~\cite{xiang2008learning}, and cosine~\cite{chen2020simple}, all have limitations under domain shifts. Euclidean distance measures the straight-line distance between feature vectors, making it sensitive to scale and distribution variations. Mahalanobis distance considers feature correlations via the covariance matrix, but remains vulnerable to domain shifts and scaling effects. Cosine similarity alleviates magnitude sensitivity but does not explicitly preserve embedding geometry when feature distributions change.
Different from the above distance metric, the geodesic distance~\cite{shamai2017geodesic} on the hypersphere preserves the angular structure of the feature space, which is more suitable for modeling domain shifts that often manifest as style variations rather than magnitude changes.
Therefore, this paper calculates the geodesic distance between samples and their corresponding class centroid in the latent metric space, which is formalized as follows:
\begin{equation}
    d_{\psi}(\mathbf{\mu}_{c}, \phi_\Theta(\mathbf{x}_{i}^{(c)})) = \frac{1}{\pi} \cdot \mathcal{G} ( \text{Proj}_{S^{n-1}}(\mathbf{\mu}_{c}) , \text{Proj}_{S^{n-1}}\phi_\Theta(\mathbf{x}_{i}^{(c)})  ), 
\end{equation}
where $\phi_\Theta(\cdot)$ is the feature extractor, $\mathcal{G}(\cdot, \cdot)$ denotes the spherical geodesic distance, $\text{Proj}_{S^{n-1}}(\mathbf{v}) = \frac{\mathbf{v}}{\left\| \mathbf{v} \right\|_2}$ indicates the spherical projection operator onto the unit hypersphere $S^{n-1} \subset \mathbb{R}^n$, and $\mathbf{\mu}_{c}$ represents the centroid of class $c$, defined as $\mathbf{\mu}_{c} = \frac{1}{N_c} \sum_{i=1}^{N_c} \phi_\Theta(\mathbf{x}_i$).

\subsection{Motivation of Our \textbf{CenterPolar}}
\label{sub:mot}
\subsubsection{\textbf{C$^3$E}}
Current SDG-DML methods rely on domain expansion-equalization strategies (\textit{e.g.}, SEE~\cite{yan2023learning}), which often suffer from inadequate diversity of expanded data, as illustrated in the top left of Fig.~\ref{fig:fig1}(a). Specifically, due to the low reliability of the class proxy, these methods typically generate augmented samples that remain overly clustered near the class proxies within the source domain distribution. This behavior leads to a limited simulation of the widespread and potentially distant domain shifts encountered in practical, unseen target domains, as shown in the top right of Fig.~\ref{fig:fig1}(a). Consequently, the learned metric model lacks sufficient exposure to the feature variations induced by significant domain changes, hindering its ability to generalize robustly across diverse unseen domains. In response to this issue, we aim to dynamically broaden the scope of the source domain distribution to encompass a wider spectrum of the potential unseen domains. Therefore, as shown in the bottom left of Fig.~\ref{fig:fig1}(a), we propose the Class-Centric Centrifugal Expansion (\textbf{C$^3$E}) phase, which intentionally shifts augmented samples away from their class centroids using centrifugal expansion, thereby actively exploring and incorporating more diverse domain characteristics into the training process.

\subsubsection{\textbf{C$^4$}}
While expanding the domain distribution is crucial for unseen domain generalization, the centrifugal expansion in \textbf{C$^3$E} inherently risks weakening the discriminative power of the learned metric space for unseen categories. Specifically, pushing samples away from their class centroids can increase intra-class variance and potentially reduce the margin between different classes in the feature space. This behavior leads to less compact intra-class clusters and weaker inter-class separation, making the model vulnerable to category shifts where distinguishing features for unseen categories might be obscured. In response to this issue, we aim to consolidate and strengthen the domain-invariant class-discriminative information essential for generalization to unseen categories. Therefore, we introduce the Class-Centric Centripetal Constraint (\textbf{C$^4$}) phase, which pulls all samples (both original and \textbf{C$^3$E}-augmented) towards their class centroids while enforcing separation between centroids, thereby ensuring high intra-class compactness and clear inter-class boundaries in the final learned metric space, as shown in the bottom right of Fig.~\ref{fig:fig1}(a).

\begin{figure*}[t] 
    \centering
    \includegraphics[width=\linewidth]{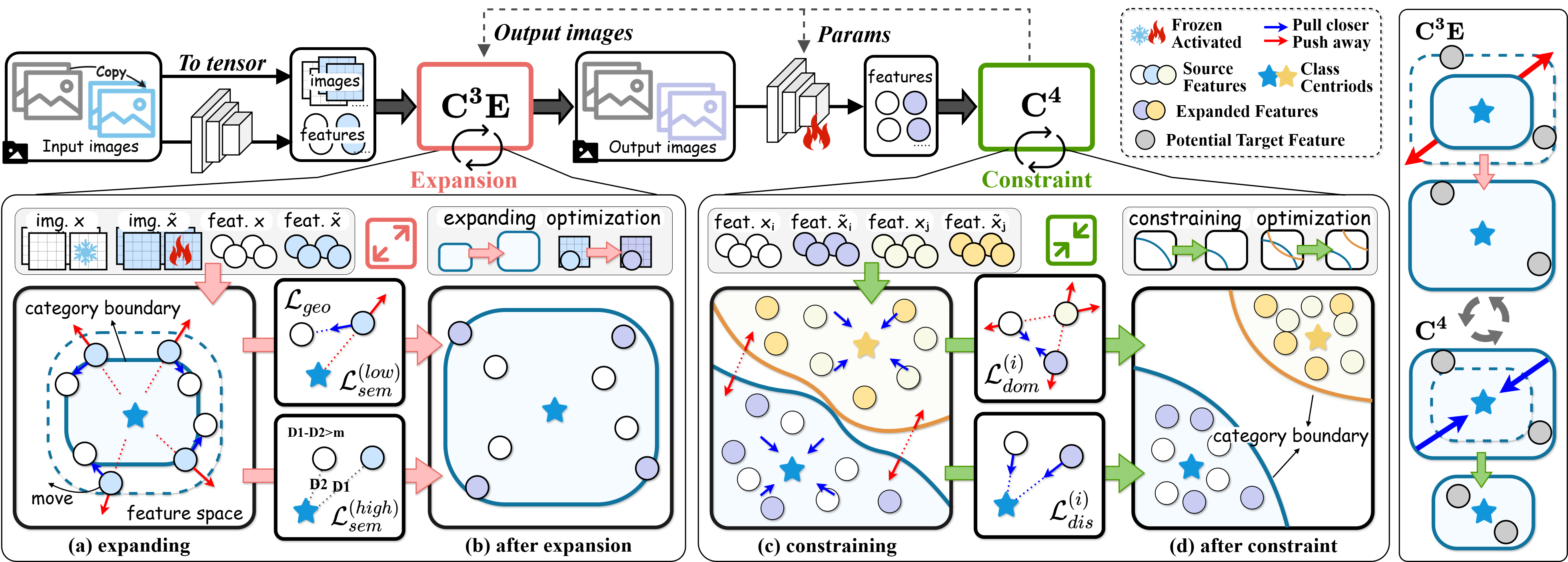}
    \caption{
    An overview of the \textbf{CenterPolar} framework, which consists of two phases: \textbf{C$^3$E} (\textit{i.e.}, Class-Centric Centrifugal Expansion) and \textbf{C$^4$} (\textit{i.e.}, Class-Centric Centripetal Constraint). \textbf{C$^3$E} expands the source domain by pushing samples away from class centroids while preserving semantic information; \textbf{C$^4$} pulls all samples toward class centroids and enforces intra-class compactness and clearer inter-class separation. 
    }
    \vspace{-3mm}
    \label{fig:framework}
\end{figure*}  

\subsection{Class-Centric Centrifugal Expansion}
\label{subsec:c3e}
We introduce the \textbf{C}$^3$\textbf{E} phase to centrifugally expand the out-of-distribution samples, providing diverse data distributions for the metric model to learn domain-invariant metric capabilities for unseen domains via \textit{Adversarial Manifold Expansion} and \textit{Semantic Boundary Preservation}.


\noindent \textbf{Adversarial Manifold Expansion.}
To begin with, we initialize the expanded sample $\tilde{\mathbf{x}}_{i(0)}$ as an identical copy of the source input sample $\mathbf{x}_i$. The pixels of $\tilde{\mathbf{x}}_{i}$ are treated as trainable parameters while freezing the class centroid $\mathbf{\mu}_{c}$ and the network parameters $\Theta$.
Specifically, we iteratively update $\tilde{\mathbf{x}}_{i}$ to maximize the geodesic distance between it and the class centroid $\mathbf{\mu}_{c}$ to enable it to exhibit new properties compared to $\mathbf{x}_{i}$. The specific loss function is defined as:
\begin{equation}
\mathcal{L}_{geo} (\tilde{\mathbf{x}}_{i}; \Theta) = -d_{\psi}(\mathbf{\mu}_{c}, \phi_\Theta(\tilde{\mathbf{x}}_{i}^{(c)})).
\label{eq:geo}
\end{equation}
\noindent \textbf{Semantic Boundary Preservation.}
On the one hand, we hope that $\tilde{\mathbf{x}}_{i}$ preserves as much of the low-level pixel semantic information as possible from $\mathbf{x}_{i}$ to avoid over-expansion. This ensures that the network can effectively learn the generalized distance metric in a progressive manner. The detailed calculation process is as follows:
\begin{equation}
\begin{aligned}
\mathcal{L}^{(low)}_{sem} (\tilde{\mathbf{x}}_{i}; \Theta) & = \inf _{\gamma \in \Gamma} \int_0^1 g_{\gamma(t)}(\dot{\gamma}(t), \dot{\gamma}(t)) d t \\
& \text { s.t. } \quad \gamma(0)=\mathbf{x}_i, \gamma(1)=\tilde{\mathbf{x}}_i.
\end{aligned}
\end{equation}
In this formulation,  $\gamma(t)$ denotes the centrifugal expansion trajectory; $\dot{\gamma}(t)$ governs both the direction and magnitude of the domain shift occurring during the centrifugal expansion process; $g_{\gamma(t)}$ characterizes the pixel-space topology (with Euclidean topology adopted by default in this work); $\Gamma$ imposes constraints on semantically plausible augmentations; and $\inf$ denotes the Infimum that ensures deformation with optimal energy efficiency.

On the other hand, in order to prevent insufficiently expanded samples from being too similar to the original input samples, thereby ensuring that the network can sufficiently learn the generalized distance metric, we expect the expanded sample $\tilde{\mathbf{x}}_{i}^{(c)}$ to maintain at least a margin $m$ away from its class centroid $\mu_{c}$ at the high-level feature semantic level. The detailed formula is as follows:
\begin{equation}
\mathcal{L}^{(high)}_{sem} (\tilde{\mathbf{x}}_{i}; \Theta) = [ d_{\rho}(\mathbf{\mu}_{c}, \phi_\Theta(\tilde{\mathbf{x}}_{i}^{(c)})) - d_{\rho}(\mathbf{\mu}_{c}, \phi_\Theta(\mathbf{x}_{i}^{(c)})) + m ]_+,
\end{equation}
where $d_{\rho}$ means Euclidean distance, $[\cdot]_+$ implies a non-negative operation, and $m$ is a score margin.

\noindent \textbf{Overall Objective for C$^3$E.} 
Finally, we can derive the overall objective of the \textbf{C}$^3$\textbf{E} phase as follows:
\begin{equation}
    \mathcal{L}_{\text{\textbf{C}}^3\text{\textbf{E}}} = \frac{1}{N_s} \sum_{i=1}^{N_s} (\mathcal{L}_{geo} (\tilde{\mathbf{x}}_{i}; \Theta) + \mathcal{L}^{(low)}_{sem} (\tilde{\mathbf{x}}_{i}; \Theta) + \mathcal{L}^{(high)}_{sem} (\tilde{\mathbf{x}}_{i}; \Theta)).
\label{eq:c3e}
\end{equation}
Furthermore, after a series of iterations through $\mathcal{L}_{\text{\textbf{C}}^3\text{\textbf{E}}}$, we can get the final expanded sample $\tilde{\mathbf{x}}_{i(*)}$ of $\mathbf{x}_i$:
\begin{equation}
    \tilde{\mathbf{x}}_{i(*)}=\underset{\tilde{\mathbf{x}}_i}{\operatorname{argmin}} \mathcal{L}_{\text{\textbf{C}}^3\text{\textbf{E}}}
\end{equation}

\subsection{Class-Centric Centripetal Constraint}
\label{subsec:c4}
After the \textbf{C}$^3$\textbf{E} phase, we use the expanded samples and original samples to further refine the learning in the \textbf{C}$^4$ phase. Specifically, we use these expanded samples and original samples to reinforce the metric model learning through \textit{Domain-Invariant Feature Consolidation} and \textit{Discriminative Metric Refinement}, which consolidates its class-discriminative generalization ability for unseen categories. The \textbf{C}$^4$ phase consolidates domain-invariant class discriminability through dual cooperative parts as follows:

\noindent \textbf{Domain-Invariant Feature Consolidation.}
To further alleviate the negative impact of domain shift, we combine the original samples $\mathcal{D}_{S} = \left\{\left(\mathbf{x}_{i}^{s}, y_{i}\right)\right\}_{i=1}^{N_s}$ and the expanded samples $\tilde{\mathcal{D}}_{S} = \left\{\left(\tilde{\mathbf{x}}_{i}^{s}, y_{i}\right)\right\}_{i=1}^{N_s}$ into a new training set $\Omega = \mathcal{D}_S\cup \tilde{\mathcal{D}}_S$, enabling the consolidation of domain-invariant features. For an input sample $(\mathbf{x}_{i}, y_i) \in \Omega$, the concrete optimization constraint in the feature space is as follows:
\begin{equation}
\begin{aligned}
    \mathcal{L}_{dom}^{(i)} = & \frac{1}{|\mathcal{P}|} \sum_{i=j} \mathcal{I}(i, j) [d_{\rho}(\phi_\Theta(\mathbf{x}_{i}), \phi_\Theta(\mathbf{x}_{j})) - m_p]_+ \\
     & + \frac{1}{|\mathcal{N}|} \sum_{i \neq j} \mathcal{I}(i, j) [m_n - d_{\rho}(\phi_\Theta(\mathbf{x}_{i}), \phi_\Theta(\mathbf{x}_{j}))]_+,
\end{aligned}
\label{eq:dom}
\end{equation}
where $m_p = 0$ and $m_n = 1$ are the positive and negative margins, respectively. $d_{\rho}$ means Euclidean distance, $[\cdot]_+$ implies a non-negative operation, and $\mathcal{P}$ and $\mathcal{N}$ correspond to the number of positive and negative samples for sample $\mathbf{x}_{i}$. Note that $\mathcal{I}(i, j)$ represents indicator function as follows:
\begin{equation}
	\mathcal{I}(i, j)= \begin{cases} 1 & i = j \\ 0 & i \neq j \end{cases}
	\label{eq:indicator}
\end{equation}
In this case, we can guarantee that the domain-invariant metric model can better capture class-discriminative ability, \textit{i.e.}, inter-class separation, across different domains.


    
    

\noindent \textbf{Discriminative Metric Refinement.}
Additionally, to further strengthen the intra-class compactness, we apply the opposite constraint to the Adversarial Manifold Expansion in \textbf{C}$^3$\textbf{E}, formalized as follows:
\begin{equation}
\mathcal{L}_{dis}^{(i)} (\mathbf{x}_{i}; \Theta) = d_{\psi}(\mathbf{\mu}_{c}, \phi_\Theta(\mathbf{x}_{i}^{(c)})),
\label{eq:dis}
\end{equation}
where $\mu_c$ is the class centroid of class $c$ in original samples $\mathcal{D}_S$, and $d_{\psi}$ denotes the geodesic distance. In particular, these expanded samples can also be regarded as hard samples in the feature space, which provides a further assessment of the generalized class-discriminative ability of the metric model for unseen classes.

\noindent \textbf{Overall Objective for C$^4$.} 
In summary, we can express the overall objective for \textbf{C}$^4$ as follows:
\begin{equation}
    \mathcal{L}_{\text{\textbf{C}}^4} = \frac{1}{2N_s} \sum_{i=1}^{2N_s} (\mathcal{L}_{dom}^{(i)} + \lambda \mathcal{L}_{dis}^{(i)}).
    \label{eq:c4}
\end{equation}

\subsection{Overall Procedure for \textbf{CenterPolar}}
\label{sec: overall}
\textbf{CenterPolar} operates through iterative alternation between two collaborative phases, \textit{i.e.}, \textbf{C$^3$E} and \textbf{C$^4$}, as illustrated in Fig.~\ref{fig:framework}. 
For \textbf{C$^3$E} phase: During the first $\mathcal{E}_e$ epochs, for each training batch, a copy of each original sample $\mathbf{x}_i$, \textit{i.e.}, $\tilde{\mathbf{x}}_{i(0)}$ is used as the initialization for the corresponding expanded sample. Then, each expanded sample $\tilde{\mathbf{x}}_i^{(c)}$ is dynamically pushed away from its class centroid $\mu_c$ via adversarial manifold expansion at the pixel level, generating expanded domain distribution data while constraining semantic shift through low-level pixel expansion and high-level feature preservation via semantic boundary preservation. 
This mitigates excessive dispersion, thereby preserving the semantic integrity of the representations.
The detailed optimization objectives of the \textbf{C$^3$E} phase are formalized in Eqs. \ref{eq:geo} to \ref{eq:c3e}.
For \textbf{C$^4$} phase: Both original and expanded samples, $\Omega = \mathcal{D}_S \cup \tilde{\mathcal{D}}_S$, are simultaneously pulled toward their class centroids in the feature space, optimizing a joint objective of domain-invariant feature consolidation (through distance margin constraints) in Eq.~\ref{eq:dom} and discriminative metric refinement (via geodesic centroid attraction) in Eq.~\ref{eq:dis}. Hence, the final optimization objective of the \textbf{C$^4$} phase is formed in Eq.~\ref{eq:c4}.
The detailed training process of our proposed \textbf{CenterPolar} framework can be referred to as Algorithm~\ref{alg:CenterPolar}.

\begin{algorithm}[t]
    \caption{Training Process with CenterPolar}
    \renewcommand{\algorithmicrequire}{\textbf{Input:}}
    \renewcommand{\algorithmicensure}{\textbf{Output:}}
    \label{alg:CenterPolar}
    \begin{algorithmic}[1]
        \REQUIRE Source domain dataset $\mathcal{D}_{S} = \{(\mathbf{x}_{i}^{s}, y_{i})\}_{i=1}^{N_s}$; Pre-trained model $\Theta$; Number of training epochs $E$; Expansion epochs $\mathcal{E}_e$; Number of expansion iterations $T_e$
        \ENSURE Final model parameters $\Theta$
        
        \STATE Initialize class centers $\boldsymbol{\mu}_{c}$ by computing the mean of class features
        \STATE Let $\mathbf{x}_o \gets \mathbf{x}$ \COMMENT{Backup original inputs}

        \FOR{$e \gets 1$ \TO $E$}
            \IF{$e \in \mathcal{E}_e$}
                \FORALL{mini-batch $(\mathbf{x}, y)$ in $\mathcal{D}_S$}
                    \STATE $\tilde{\mathbf{x}} \gets \mathbf{x}_o$ \COMMENT{Start from original inputs}
                    \FOR{$t \gets 1$ \TO $T_e$}
                        \STATE Compute $\mathcal{L}_{\mathbf{C}^3\mathbf{E}}$ using Eq.~\ref{eq:c3e}
                        \STATE $\tilde{\mathbf{x}} \gets \tilde{\mathbf{x}} - \nabla_{\tilde{\mathbf{x}}} \mathcal{L}_{\mathbf{C}^3\mathbf{E}}$
                    \ENDFOR
                    \STATE Add augmented pairs $(\tilde{\mathbf{x}}, y)$ into $\tilde{\mathcal{D}}_{S}$
                    \STATE $\Omega \gets \mathcal{D}_{S} \cup \tilde{\mathcal{D}}_{S}$
                    \STATE $\mathbf{x}_o \gets \tilde{\mathbf{x}}$
                \ENDFOR
            \ENDIF
            \FORALL{mini-batch $(\mathbf{x}, \mathbf{y})$ in $\Omega$}
                \STATE Compute $\mathcal{L}_{\mathbf{C}^4}$ using Eq.~\ref{eq:c4}
                \STATE $\Theta \gets \Theta - \nabla_{\Theta} \mathcal{L}_{\mathbf{C}^4}$
            \ENDFOR
        \ENDFOR

        \STATE \textbf{return} $\Theta$
    \end{algorithmic}
\end{algorithm}
\vspace{-2mm}

\subsection{Theoretical Analysis of \textbf{CenterPolar}}
\label{sub:theor}   


\subsubsection{\textbf{Bounded Centrifugal Expansion}}
The \textbf{C$^3$E} phase operates by optimizing the expanded sample $\tilde{\mathbf{x}}$ to maximize its pixel-level geodesic separation from the class centroid \(\mu_c\). This centrifugal constraint is counteracted by a high-level hinge margin $m$, which explicitly restricts the drift to prevent mode collapse (refer to Eq.~\ref{eq:c3e} in \S~\ref{subsec:c3e}). In the \textbf{C$^3$E} phase, the complete gradient of the expanded sample \(\tilde{\mathbf{x}}\) is:


\begin{equation}
\nabla_{\tilde{\mathbf{x}}}\mathcal{L}_{\text{C}^3\text{E}} =  -\frac{\partial d_{\psi}}{\partial \phi}\bigg|_{\phi=\phi_\Theta(\tilde{\mathbf{x}})} \frac{\partial \phi_\Theta(\tilde{\mathbf{x}})}{\partial \tilde{\mathbf{x}}} + \nabla_{\tilde{\mathbf{x}}} \mathcal{L}^{(low)}_{sem} + \mathbb{I}[\mathcal{L}^{(high)}_{sem}>0] \cdot \frac{\partial d_{\rho}}{\partial \phi}\bigg|_{\phi=\phi_\Theta(\tilde{\mathbf{x}})} \frac{\partial \phi_\Theta(\tilde{\mathbf{x}})}{\partial \tilde{\mathbf{x}}},
\end{equation}
where \(\mathbb{I}[\cdot]\) is an indicator function, which takes a value of $1$ when the boundary constraints are in effect. The first term expands the local manifold by increasing the geodesic distance to the class centroid \(\mu_c\); the second term preserves low-level semantic information with the help of an optimal transport constraint; and the hinge term introduces a stopping mechanism: when \(d_\rho(\mu_c, \phi_\Theta(\tilde{\mathbf{x}})) \ge d_\rho(\mu_c, \phi_\Theta(\mathbf{x})) + m\) in the high-level semantic feature space, the indicator function \(\mathbb{I}[\cdot]\) takes on the value zero, at which point the expansion gradient is offset by the gradient of the semantically preserved term. Therefore, the expansion process is bounded by both the margin \(m\) and the semantic constraint.

\subsubsection{\textbf{Controlled Centripetal Constraint}}

The \textbf{C$^4$} phase optimizes the learned feature representation by shrinking the features toward their corresponding class centroid. At the same time, domain invariance constraints are used to maintain inter-class separability (refer to Eq.~\ref{eq:c4} in \S~\ref{subsec:c4}). The full gradient of the network parameter \(\Theta\) is as follows:

\begin{equation}
\nabla_{\Theta}\mathcal{L}_{\text{C}^4} = \frac{1}{2N_s} \sum_{i=1}^{2N_s} \nabla_{\Theta} \mathcal{L}_{dom}^{(i)} + \frac{\lambda}{2N_s} \sum_{i=1}^{2N_s} \frac{\partial d_{\psi}}{\partial \phi}\bigg|_{\phi=\phi_\Theta(\mathbf{x}_i)} \frac{\partial \phi_\Theta(\mathbf{x}_i)}{\partial \Theta}.
\end{equation}


During gradient descent, the second term achieves controlled constraint in the feature space by shortening the geodesic distance between samples and their corresponding class centroids. The first term (\(\nabla_{\Theta} \mathcal{L}_{dom}^{(i)}\)) ensures inter-class separability by bringing positive pairs closer together and negative pairs further apart. This constraint is inherently controllable: by choosing an appropriate value of \(\lambda\), we can avoid collapse of the feature space while enhancing the compactness of intra-class features.  

\subsubsection{\textbf{Two Lemmas}}
\textbf{Lemma 1 (Bounded Expansion).}  
Given the \textbf{C$^3$E} objective with semantic constraints and hinge margin \(m\), 
the direction of the hinge term gradient is reversed and cancels out the dilation gradient:
\begin{equation}
\nabla_{\tilde{\mathbf{x}}}\mathcal{L}_{\text{C}^3\text{E}} \approx -\frac{\partial d_{\psi}}{\partial \phi}\bigg|_{\phi=\phi_\Theta(\tilde{\mathbf{x}})} \frac{\partial \phi_\Theta(\tilde{\mathbf{x}})}{\partial \tilde{\mathbf{x}}} + \nabla_{\tilde{\mathbf{x}}} \mathcal{L}^{(low)}_{sem}.
\end{equation}
Thus, when the semantic constraint dominates, expansion ceases, ensuring bounded manifold growth.

\noindent \textbf{Lemma 2 (Controlled Constraint).}  
The \textbf{C$^4$} objective comprises two opposing forces: a centripetal shrinkage (weighted by $\lambda$) and a discriminative separation mechanism. At convergence, their gradients are counterbalanced, resulting in a stable equilibrium:
\begin{equation}
\frac{1}{2N_s} \sum_{i=1}^{2N_s} \nabla_{\Theta} \mathcal{L}_{dom}^{(i)} + \frac{\lambda}{2N_s} \sum_{i=1}^{2N_s} \nabla_{\Theta} d_{\psi}(\mu_c,\phi_\Theta(\mathbf{x}_i)) \approx 0.
\end{equation}
Thus, the shrinkage strength can be adjusted via \(\lambda\): a smaller \(\lambda\) results in gentle merging while preserving intra-class boundaries, while a larger \(\lambda\) may lead to excessive shrinkage and loss of discriminative capability.


\subsubsection{\textbf{Gradient Dynamic Mechanism and Optimization Stability}}

In summary, \textbf{C$^3$E} primarily acts on the input sample \(\tilde{\mathbf{x}}\), using gradients to push expanded samples away from their class centroids along the geodesic direction (\emph{expansion}), while using semantic constraints to prevent excessive deformation of sample features.
In contrast, \textbf{C$^4$} acts on the network parameters \(\Theta\), pulling the generated features toward their class centroids (\emph{constraint}),
while maintaining intra-class separation. The margin \(m\) and weight \(\lambda\) act as complementary control factors: \(m\) limits the upper bound of feature space expansion via the hinge activation function, while \(\lambda\) controls the strength of constraint via the gradient balancing function. Together, they form a stable push-pull dynamic that broadens domain coverage while preserving class distinction.

\section{Experiments}
\label{sec:experiments}

In this section, we first introduce the experimental setup in \S~\ref{sub:set}. Second, we compare our method with a number of state-of-the-art DML and SDG-DML methods in \S~\ref{sub:comp} to evaluate the generalization ability of different methods and provide analysis. Third, we present an ablation study in \S~\ref{sub:abl}. Fourth, we present visualization results in \S~\ref{sub:vis}. Finally, we show the computational complexity and limitations of our method in \S~\ref{sub:time} and \S~\ref{sub:lim}, respectively.


\subsection{Experimental Settings}
\label{sub:set}
\noindent \textbf{Datasets and Evaluation Metrics.} We conduct experiments on CUB-200-2011~\cite{wah2011caltech}, Cars196~\cite{krause20133d}, DomainNet~\cite{peng2019moment}, PACS~\cite{li2017deeper}, and Office-Home~\cite{venkateswara2017deep}. These datasets collectively cover a broad spectrum of fine-grained recognition (CUB, Cars) and multi-domain object recognition (DomainNet, PACS, Office-Home), featuring diverse visual styles, varying sample sizes, and inherent class imbalance, thus naturally aligning with the challenging SDG-DML scenario. 
In addition to style shift, they also exhibit differences in viewpoint, background, and illumination, which further increase the difficulty of generalization.
However, both CUB-200-2011 and Cars196 are single-style datasets and therefore cannot be directly applied to the SDG-DML setting. Therefore,
following the settings~\cite{yan2023learning}, we use StyleNet~\cite{zou2021stylized} to generate CUB-200-2011 Ext. and Cars196 Ext.. Table~\ref{tab:datasets statistics} shows more details about the aforementioned datasets, training, and testing split settings. We abbreviate some words into signs for a more concise display and list the words corresponding to the signs in Table~\ref{tab:description}.
Evaluation metrics~\cite{musgrave2020metric}: Recall@1, Recall@2, RP, and MAP@R. Recall@$k$ measures retrieval accuracy in the top-$k$, RP assesses relevance, and MAP@R reflects overall ranking quality. 

\begin{table}[t]
\centering
\caption{The training/testing statistics of seven datasets: CUB-200-2011, CUB-200-2011 Ext., Cars196, Cars196 Ext., DomainNet, PACS, and Office-Home.}
\renewcommand{\thetable}{I}
\resizebox{0.6\linewidth}{!}{
\begin{tabular}{l||ccc}
\Xhline{1.5pt}
& \multicolumn{3}{c}{\textbf{Training / Tesing}} \\
\cline{2-4}
\multirow{-2}{*}{\textbf{Dataset}} & \# Samples & \# Classes & Domains\\
\hline
\hline
CUB-200-2011 & 5,864 / 5,924 & 100 / 100 & 1 / 1 \\
CUB-200-2011 Ext. & 5,864 / 5,924 & 100 / 100 & 1 / 2 \\
Cars196 & 8,054 / 8,131 & 98 / 98 & 1 / 1 \\
Cars196 Ext. & 8,054 / 8,131 & 98 / 98 & 1 / 2 \\
DomainNet & 89,622 / 224,525 & 173 / 172 & 1 / 5 \\
PACS & 759 / 3,018 & 4 / 3 &  1 / 3\\
office-Home & 2,358 / 5,157 & 33 / 32 &  1 / 3\\
\Xhline{1.5pt}
\end{tabular}}
\label{tab:datasets statistics}
\end{table}

\begin{table}[t]
\centering
\caption{Description of the corresponding signs in the text.}
\resizebox{0.55\linewidth}{!}{
\begin{tabular}{c|l|c}
 \Xhline{1.5pt}
 \textbf{Dataset} & \textbf{Sign}  & \textbf{Description} \\
\hline
- & \textbf{$\rightarrow$} & Domain shift from training to testing. \\
\hline
\multirow{2}{*}{CUB Ext.} &   R $\rightarrow$ O & Real Image $\rightarrow$ Oil-painting \\
& R $\rightarrow$ W &  Real Image $\rightarrow$ Water-painting\\
\hline
\multirow{2}{*}{Cars Ext.} &   R $\rightarrow$ O & Real Image $\rightarrow$ Oil-painting \\
& R $\rightarrow$ W &  Real Image $\rightarrow$ Water-painting\\
\hline
\multirow{5}{*}{Domain} & R $\rightarrow$ S & Real $\rightarrow$ Sketch \\
&R $\rightarrow$ I & Real $\rightarrow$ Infograph \\
&R $\rightarrow$ P & Real $\rightarrow$ Painting \\
&R $\rightarrow$ Q & Real $\rightarrow$ Quickdraw \\
&R $\rightarrow$ C & Real $\rightarrow$ Clipart \\
\hline
\multirow{3}{*}{PACS} & P $\rightarrow$ A & Photo $\rightarrow$ Art painting \\
&P $\rightarrow$ C & Photo $\rightarrow$ Cartoon \\
&P $\rightarrow$ S & Photo $\rightarrow$ Sketch \\
\hline
\multirow{3}{*}{Office.} & R $\rightarrow$ A & Real World $\rightarrow$ Art\\
&R $\rightarrow$ C & Real World $\rightarrow$ Clipart \\
&R $\rightarrow$ P & Real World $\rightarrow$ Product \\
\hline
-& MAP & MAP@R \\
 \Xhline{1.5pt} 
\end{tabular}
}
\vspace{-4mm}
\label{tab:description}
\end{table}

\noindent \textbf{Image Preprocessing.} To ensure a fair comparison, we preprocess the images following the procedure used in most existing DML research~\cite{kim2020proxy,wang2019multi}. For the training and testing sets of CUB-200-2011 Ext., DomainNet, PACS, Office-Home, we resize all images to 256 pixels, and then perform standard center cropping to adjust them to 224 $\times$ 224. For Cars196 Ext., since resizing to 256 would damage the main subject information, we first resize the training and testing sets to 256 $\times$ 256, and then perform standard center cropping to adjust them to 224 $\times$ 224.

\noindent \textbf{Backbone and Parameters.} Experiments are conducted on PyTorch using ResNet-50~\cite{he2016deep} pretrained on ImageNet. Note that we use contrastive loss~\cite{hadsell2006dimensionality} as the baseline. SGD expands pixels, while Adam updates network parameters. For each iteration, we set the batch size to 32 for Cars196 Ext., to 64 for CUB-200-2011 Ext., and to 256 for DomainNet. The embedding size is fixed at 512. A full training cycle comprises 40 rounds for all datasets, with testing conducted every two rounds. CUB-200-2011 Ext.: C$^3$E is performed every 15 rounds, starting from round 1, for a total of 20 optimization rounds. Cars196 Ext.: C$^3$E occurs every 15 rounds, starting from round 1, for a total of 5 optimization rounds. DomainNet \& PACS \& Office-Home: C$^3$E is applied every 20 rounds, starting from round 1, for a total of 10 optimization rounds. We set the learning rate for all two-stage networks to $1 \times 10^{-6}$. Additionally, we set \(m\) to 1; on CUB-200-2011 Ext., we set $\lambda$ to 0.75, on Cars196 Ext., we set it to 0.0001, and on DomainNet, we set it to 0.001; on PACS and Office-Home, we set $\lambda$ to 1. 
All experimental results were conducted with an NVIDIA RTX 4080 SUPER GPU (32GB).


\subsection{Comparison with State-of-the-art Methods}
\label{sub:comp}
We compare our method with state-of-the-art DML and SDG-DML methods. DML methods include methods: Triplet loss~\cite{hoffer2015deep}, Margin loss~\cite{wu2017sampling} , ProxyNCA loss (PNCA)~\cite{movshovitz2017no}, CosFace loss~\cite{wang2018cosface}, MultiSimilarity loss (MS)~\cite{wang2019multi}, Circle loss~\cite{sun2020circle}, Proxyanchor loss (PA)~\cite{kim2020proxy}, SupCon loss~\cite{khosla2020supervised}, Instance loss~\cite{zheng2020dual}, DCML~\cite{zheng2021deep}, DAS~\cite{liu2022densely}, HIST~\cite{lim2022hypergraph}, VICReg~\cite{bardes2022vicreg}, HSE~\cite{yang2023hse}, IDML~ \cite{10239539}, DADA~\cite{ren2024towards}, AntiCo~\cite{jiang2024anti}, DDML~\cite{park2025deep}, and PDML~\cite{bhatnagar2025potential}. SDG-DML methods only contain: SEE~\cite{yan2023learning}.
We conducted comprehensive experimental evaluations on five benchmark datasets shown in Table~\ref{tb:sexperiment-results}, Table~\ref{tab:pacs+}, and Table~\ref{tab:DomainNet results}. We used two categories of methods for comparison: DML and SDG-DML methods. The following is our detailed experimental result analysis.

\begin{table*}[t]
\centering
\caption{Comparison with the state-of-the-art DML and SDG-DML methods on CUB-200-2011 Ext. \textbf{(Left)} and Cars196 Ext. \textbf{(Right)} datasets, respectively. \colorbox{graytxt}{The top part} presents DML methods, while \colorbox{bluelight}{the bottom part} shows SDG-DML methods. \textbf{Bold} and \underline{underline} indicate the best and second-best results, respectively.}
\small
\resizebox{\linewidth}{!}{
\setlength{\tabcolsep}{1.1mm}
\begin{tabular}{l || c c c c | c c c c || c c c c | c c c c}
\Xhline{1.5pt}
 & \multicolumn{8}{c||}{\textbf{CUB-200-2011 Ext.}} & \multicolumn{8}{c}{\textbf{Cars196 Ext.}} \\
\cline{2-17} 
& \multicolumn{4}{c|}{\textbf{R $\rightarrow$ O}} & \multicolumn{4}{c||}{\textbf{R $\rightarrow$ W}} & \multicolumn{4}{c|}{\textbf{R $\rightarrow$ O}} & \multicolumn{4}{c}{\textbf{R $\rightarrow$ W}} \\
\cline{2-17} 
 \multirow{-3}{*}{\textbf{Method}} & R@1 & R@2 & RP & MAP & R@1 & R@2 & RP & MAP & R@1 & R@2 & RP & MAP & R@1 & R@2 & RP & MAP\\
\hline
\hline
\cellcolor{graytxt}Triplet~\cite{hoffer2015deep} \scriptsize{[CVPR'15]} & 21.39 & 31.50 & 10.46 & 4.16 &24.00 & 34.72 & 11.34 &4.54 & 17.93 & 27.06 & 6.67  & 1.62 &18.34 & 26.68 & 6.03  & 1.44\\
\cellcolor{graytxt}Margin~\cite{wu2017sampling} \scriptsize{[ICCV'17]} & 21.74 & 31.41 & 10.87 & 4.54 & 22.79 & 32.02 & 10.94 & 4.45 & 17.98 & 26.55 & 6.51  & 1.67 & 17.49 & 25.63 & 5.39  & 1.28\\
\cellcolor{graytxt} PNCA~\cite{movshovitz2017no} \scriptsize{[ICCV'17]} & 20.32 & 29.22 & 9.61  & 3.83 & 20.58 & 29.37 & 9.67  &3.75  & 18.79 & 27.94 &6.70  & 1.77 & 18.58 & 27.28 & \underline{6.04}  &1.54\\
\cellcolor{graytxt}CosFace~\cite{wang2018cosface} \scriptsize{[CVPR'18]} & 19.72 & 28.85 & 9.35  & 3.62 & 20.70 & 30.13 & 10.26 & 4.02 & 15.05 & 23.15 & 5.40  & 1.29 & 16.31 & 23.91 & 4.91  & 1.13\\
\cellcolor{graytxt}MS~\cite{wang2019multi} \scriptsize{[CVPR'19]} & 22.21 &31.72 &10.50 & 4.30 & 24.49 & 33.93 &11.20 & 4.55 & 19.09 & \underline{28.40} & \underline{6.95}  &1.81 &18.74 & \underline{27.50} & 5.99  &1.45\\
\cellcolor{graytxt}Circle~\cite{sun2020circle} \scriptsize{[CVPR'20]}  & 20.21 & 29.27 & 9.23  & 3.45 & 22.47 & 32.44 & 10.20 & 3.80 & 17.71 & 26.71 & 6.10  & 1.49 & 16.71 & 24.68 & 5.19  & 1.15\\
\cellcolor{graytxt}PA~\cite{kim2020proxy} \scriptsize{[CVPR'20]} & 19.78 & 27.94 &9.01  & 3.39 & 20.64 & 29.88 & 9.42  & 3.48 &16.43 & 24.89 & 5.62  & 1.36 & 16.91 & 24.87 & 5.13  & 1.17\\
\cellcolor{graytxt}SupCon~\cite{khosla2020supervised} \scriptsize{[NeurIPS'20]}  & 21.57 & 31.23 & 11.25 & 4.79 & 24.46 & 33.96 & 11.60 & 4.91 & 16.05  & 23.66  & 5.85  & 1.45  & 15.41  & 22.79  & 5.16  & 1.23\\
\cellcolor{graytxt}DCML~\cite{zheng2021deep} \scriptsize{[CVPR'21]} &23.75&\underline{33.58}&\underline{11.68}&5.05 &25.20 &35.35 &11.86&5.07&15.63&23.29&5.69&1.42 &15.74&23.55&5.37&1.28\\
\cellcolor{graytxt}HIST~\cite{lim2022hypergraph} \scriptsize{[CVPR'22]} & 17.51  & 26.65  & 7.98  & 2.77  & 20.04  & 29.62  & 9.93  & 3.29 & 15.93 & 22.81 & 5.16 & 1.39 & 16.03 & 23.33 & 4.92 & 1.31\\
\cellcolor{graytxt}HSE~\cite{yang2023hse} \scriptsize{[ICCV'23]} & 19.11 &28.66 & 9.47  &3.60 & 19.14 & 28.75& 9.44   & 3.59 &13.73 & 19.71 &4.40 & 1.02 & 14.46 & 20.51 & 4.33 &1.05\\
\cellcolor{graytxt}IDML~\cite{10239539} \scriptsize{[TPAMI'24]} & 17.83  & 26.15  & 8.50  & 2.96  & 21.73  & 31.43  & 9.68  & 3.60 &  16.05  & 23.49  & 5.65  & 1.46  & 16.31  & 23.07  & 5.11  & 1.31\\
\cellcolor{graytxt}AntiCo~\cite{jiang2024anti} \scriptsize{[TMM'24]} & 22.52  & 32.63  & 10.75  & 4.44  & 24.54  & 34.25  & 11.27  & 4.76 &  \underline{19.37} & 27.55 &  6.56 &  \underline{2.00} &  18.89 & 26.91 & 5.61 & \underline{1.63}\\
\cellcolor{graytxt}DADA~\cite{ren2024towards} \scriptsize{[AAAI'24]} & 19.46  & 28.21  & 8.83  & 3.20  & 21.83  & 31.03  & 9.78  & 3.69 &  19.22 & 27.83 &  6.48 &  1.78 &  \underline{18.92} & 27.43 & 5.84 & 1.53\\
\cellcolor{graytxt}PDML~\cite{bhatnagar2025potential} \scriptsize{[CVPR'25]} & 20.39 & 30.03 & 9.19  & 3.15 & 25.00 & 35.01 & 10.71  & 4.04 & 17.34 & 25.21 & 5.63  & 1.33 & 16.19 & 23.85 & 5.17  & 1.14\\
\cellcolor{graytxt}DDML~\cite{park2025deep} \scriptsize{[AAAI'25]} &18.80 &27.14 &8.70 &3.06 &21.49&30.79&9.43&3.47 &15.80&23.11&5.59&1.41&15.64&22.78&5.16&1.31\\
\hline
\cellcolor{bluelight} SEE~\cite{yan2023learning} \scriptsize{[ICCV'23]} &  \textbf{25.71} & \textbf{35.58} & 11.67  & \underline{5.09} & \textbf{26.89} & \textbf{36.48} & \underline{12.33}  & \underline{5.33} & 15.66 &22.84 & 5.18  & 1.22 & 16.16 &23.92 & 4.88  & 1.13\\
\cellcolor{bluelight}\textbf{CenterPolar \scriptsize{[Ours]}}  & \underline{24.53} & 33.14 & \textbf{12.52} & \textbf{5.79} & \underline{26.08} & \underline{35.96} & \textbf{12.85} & \textbf{5.84} & \textbf{20.38} & \textbf{29.43} & \textbf{7.43} & \textbf{2.07} & \textbf{19.70} & \textbf{28.51} & \textbf{6.49} & \textbf{1.69}\\
\Xhline{1.5pt}
\end{tabular}     
}
\label{tb:sexperiment-results}
\vspace{-3mm}
\end{table*}

\subsubsection{\textbf{CUB-200-2011 Ext.}}
As shown in the left of Table~\ref{tb:sexperiment-results}, our proposed \textbf{CenterPolar} method surpasses existing methods across all MAP and RP metrics. On CUB-200-2011 Ext., in the Oil-painting domain, our RP and MAP achieved 12.52\% and 5.79\%, respectively, a 0.7\% improvement in MAP compared to SEE (5.79\% vs. 5.09\%). 
Among DML methods, DCML achieves the best performance, obtaining the best R@2 and RP (33.58\% and 11.68\%).
In the Water-painting domain, our RP and MAP achieved 12.85\% and 5.84\%, respectively, a 0.51\% improvement in MAP compared to SEE (5.84\% vs. 5.33\%). 
Overall, the SDG-DML method outperforms traditional DML methods, indicating the advantages of generalizing modeling styles.
However, \textbf{CenterPolar} slightly lags behind SEE in the R@1 and R@2 metrics in both domains. For example, in the watercolor domain, \textbf{CenterPolar}'s R@1 is 0.81\% lower than that of the SEE method (26.08\% vs. 26.89\%). This phenomenon indicates that \textbf{CenterPolar} favors global search ranking performance (MAP) over nearest neighbor (Top-1 / Top-2) matching.  
This trade-off arises because \textbf{CenterPolar} prioritizes global structure stability over aggressive nearest-neighbor discrimination.
On CUB Ext., the sample distribution of each class is relatively concentrated, and intra-class variation primarily stems from pose and scene variations. Therefore, the proxy well represents the class center, and SEE's proxy-based optimization process effectively aligns feature distributions across styles, achieving superior results.

\subsubsection{\textbf{Cars196 Ext.}}
As shown in the right of Table~\ref{tb:sexperiment-results}, our proposed \textbf{CenterPolar} method surpasses existing methods across all R@1, R@2, MAP, and RP metrics. On Cars196 Ext., \textbf{CenterPolar} achieved state-of-the-art results across all metrics. In the Oil-painting domain, we achieved the best R@1, R@2, RP, and MAP of 20.38\%, 29.43\%, 7.43\%, and 2.07\%, respectively. In particular, R@1 and MAP outperformed SEE by 4.72\% (20.38\% vs. 15.66\%) and 0.85\% (2.07\% vs. 1.22\%), respectively. 
Among DML methods, AntiCo achieves the best R@1 and MAP (19.37\% and 2.00\%); MS achieves the best R@2 and RP (28.40\% and 6.95\%). 
Among DML methods, DADA achieves the best R@1 (18.92\%); AntiCo achieves the best MAP (1.63\%).
Notably, SEE's performance on Cars degrades significantly, with a MAP of only 1.22\% (R $\to$ O) and 1.13\% (R $\to$ W). This phenomenon reveals the limitations of the proxy-based SEE approach in datasets with high intra-class variance (Cars). On Cars196 Ext., intra-class variance increases significantly, primarily due to variations in color, viewpoint, lighting, background, and pose. This makes it difficult for the proxy to accurately represent the overall class characteristics. When the deviation between the proxy and the true sample distribution increases, SEE's proxy-based approximate representation fails to fully capture the complex intra-class structure in high-variance data, resulting in easy overlap between different classes in the feature space and difficulty in clustering similar samples.

\subsubsection{\textbf{PACS}}
As shown in the left of Table~\ref{tab:pacs+}, we conducted a systematic evaluation on PACS. Our method achieved the best overall performance on PACS (mean R@1/RP/MAP@R: 95.35\% / 76.03\% / 67.92\%), especially MAP, which is 26.02\% higher than SEE (67.92\% vs. 41.90\%). Our method demonstrated significant advantages: in the Art painting domain, it achieved optimal performance across all four evaluation metrics (R@1: 96.08\%, R@2: 98.31\%, RP: 75.46\%, MAP@R: 67.60\%), with MAP@R improving by 36.62\% over the SEE method (67.60\% vs. 30.98\%). In the Cartoon and Sketch domains, our method maintained its lead in both RP and MAP@R, particularly in the Cartoon domain, where MAP@R is twice that of the SEE method (51.67\% vs. 24.88\%). 
Among DML methods, DCML, IDML, and PDML demonstrate strong performance. For example, DCML achieved the best average MAP (61.13\%), and PDML achieved the best R@2 (97.64\% and 98.01\%).

\begin{table*}[t]
\centering
\caption{Comparison with the state-of-the-art DML and SDG-DML methods on PACS \textbf{(Left)} and Office-Home \textbf{(Right)} datasets, respectively. \colorbox{graytxt}{The left part} corresponds to DML methods, while \colorbox{bluelight}{the right part} belongs to SDG-DML methods. \textbf{Bold} and \underline{underline} indicate the best and second-best results, respectively.
}
\resizebox{\linewidth}{!}{
\setlength{\tabcolsep}{1.1mm}
\begin{tabular}{l|l|ccccccc|cc||l|l|ccccccc|cc}
\Xhline{1.5pt}
\multicolumn{2}{c|}{\rotatebox{-90}{\small \textbf{Method}}} &  \cellcolor{graytxt}\rotatebox{-90}{\scriptsize\textbf{DCML~\cite{zheng2021deep}}\scriptsize{[CVPR'21]}} & \cellcolor{graytxt}\rotatebox{-90}{\scriptsize\textbf{HIST~\cite{lim2022hypergraph}}\scriptsize{[CVPR'22]}}  &  \cellcolor{graytxt}\rotatebox{-90}{\scriptsize\textbf{HSE~\cite{yang2023hse}} \scriptsize{[ICCV'23]}}& \cellcolor{graytxt}\rotatebox{-90}{\scriptsize\textbf{IDML~\cite{10239539}} \scriptsize{[TPAMI'24]}}& 
\cellcolor{graytxt}\rotatebox{-90}{\scriptsize\textbf{DADA~\cite{ren2024towards}} \scriptsize{[AAAI'24]}}& 
\cellcolor{graytxt}\rotatebox{-90}{\scriptsize\textbf{PDML~\cite{bhatnagar2025potential}} \scriptsize{[CVPR'25]}} & 
\cellcolor{graytxt}\rotatebox{-90}{\scriptsize\textbf{DDML~\cite{park2025deep}} \scriptsize{[AAAI'25]}} &  \cellcolor{bluelight}\rotatebox{-90}{\scriptsize\textbf{SEE~\cite{yan2023learning}} \scriptsize{[ICCV'23]}} &  \cellcolor{bluelight}\rotatebox{-90}{\scriptsize\textbf{CenterPolar (Ours)}} 
& \multicolumn{2}{c|}{ \rotatebox{-90}{\small \textbf{Method}}}  &   \cellcolor{graytxt}\rotatebox{-90}{\scriptsize\textbf{DCML~\cite{zheng2021deep}} \scriptsize{[CVPR'21]}} & \cellcolor{graytxt}\rotatebox{-90}{\scriptsize\textbf{HIST~\cite{lim2022hypergraph}} \scriptsize{[CVPR'22]}}  &  \cellcolor{graytxt}\rotatebox{-90}{\scriptsize\textbf{HSE~\cite{yang2023hse}} \scriptsize{[ICCV'23]} }&\cellcolor{graytxt} \rotatebox{-90}{\scriptsize\textbf{IDML~\cite{10239539}} \scriptsize{[TPAMI'24]}}& \cellcolor{graytxt}\rotatebox{-90}{\scriptsize\textbf{DADA~\cite{ren2024towards}} \scriptsize{[AAAI'24]}}& 
\cellcolor{graytxt}\rotatebox{-90}{\scriptsize\textbf{PDML~\cite{bhatnagar2025potential}} \scriptsize{[CVPR'25]}} & 
\cellcolor{graytxt}\rotatebox{-90}{\scriptsize\textbf{DDML~\cite{park2025deep}} \scriptsize{[AAAI'25]}} &  \cellcolor{bluelight}\rotatebox{-90}{\scriptsize\textbf{SEE~\cite{yan2023learning}} \scriptsize{[ICCV'23]}} &  \cellcolor{bluelight}\rotatebox{-90}{\scriptsize\textbf{CenterPolar (Ours)}} \\
\hline
\hline
\multirow{4}{*}{\rotatebox{90}{ \small \textbf{P$\to$A}}} & \small R@1 &94.39 &94.60& 94.92 &95.03 &\underline{95.13} &\underline{95.13} &94.71 &67.09 & \textbf{96.08} & \multirow{4}{*}{\rotatebox{90}{ \small \textbf{R$\to$A}}} & \small  R@1 &69.25 &\underline{71.47} & 71.05 & 69.88 &71.37 &69.78 & \underline{71.47}& 45.28 & \textbf{72.11} \\
 &  \small R@2  &97.57 & 97.77 & 97.35 & 97.35 &\underline{98.10} &97.35 & 97.78& 81.16 & \textbf{98.31}  &  & \small  R@2 &77.41  &80.91 & 79.43 & 79.53 & 79.75 &77.73  &\underline{81.12}& 52.92  &\textbf{81.34} \\
 &  \small RP &67.01  &66.96 & 65.19 & \underline{68.08} &65.73 &65.99 &67.04 & 50.13 & \textbf{75.46
 }&  &  \small RP &\textbf{38.88}  & 37.44 & 36.91 & 36.75 &37.64 &36.22  &37.39& 21.64 & \underline{38.78}   \\
 &  \small MAP  &56.85 & 57.35& 54.84 & \underline{58.76} &55.68 &55.50 &57.45 &30.98 & \textbf{67.60} & &  \small MAP &\underline{29.31}  & 27.81 & 27.27 &26.91&27.45 &26.36  &27.77&13.31 & \textbf{29.43} \\
\hline
\multirow{4}{*}{\rotatebox{90}{ \small \textbf{P$\to$C}}} &  \small R@1  &92.23  &\textbf{93.31}  & 92.43 & \underline{92.92}&91.05 &92.72 &92.13  & 77.09 & 92.43 & \multirow{4}{*}{\rotatebox{90}{ \small \textbf{R$\to$C}}}  \small & R@1 &\textbf{77.79}  & 76.57 & 75.82 & 76.39 &76.39 &\underline{77.75}  &76.67 & 64.94 & 77.56\\
 &  \small R@2  &96.36 & \underline{97.44}& 96.46 & 96.36 &95.97  &\textbf{97.64} &96.36 & 88.20 & 96.66 &  &  \small R@2  &\textbf{83.17} & 81.81 & 81.91 & 82.00  &82.66 &\underline{83.08} &82.09 & 71.20 &82.61\\
 & \small  RP  &\underline{63.67}  &58.88 & 58.25 & 59.35 &60.01 &59.32 & 57.68& 43.38 & \textbf{64.11} &  & \small  RP  &\textbf{28.66}  & 26.99 & 26.11 & 25.93 &26.95 &26.61   &26.93& 19.24 & \underline{27.55}\\
 &  \small MAP  &\underline{51.59} & 45.34& 44.33 & 46.10 &46.60 &45.65 &43.92 &24.88 &\textbf{ 51.67 }& &  \small MAP  &\textbf{19.02} & 17.80
 & 16.80 & 16.80 &17.42 &17.80 &17.73 & 10.56 & \underline{18.36}\\
\hline
\multirow{4}{*}{\rotatebox{90}{ \small \textbf{P$\to$S}}} & \small  R@1  &97.44  & 97.63& \textbf{98.20} & 97.92  &97.35 &\underline{98.01} &97.54 & 95.74 & 97.54 & \multirow{4}{*}{\rotatebox{90}{ \small \textbf{R$\to$P}}} &  \small R@1  &92.34 & \underline{93.68}& 93.49 & \textbf{93.88} &93.11 &93.59 &93.59  & 77.25 & 93.25\\
 &  \small R@2  &98.77 & 98.67& \underline{99.15} & \textbf{99.72} &98.86 &99.05& 98.67 & 97.82 &98.86 &  &  \small R@2  &94.41 & 95.47& \underline{95.66} & 95.42 &\underline{95.66} &\textbf{95.71 }&95.42 & 82.75 & 95.18\\
 & \small  RP  &81.13  &82.67& \underline{82.77} & 82.45 &82.34 &82.48 &83.14 & 77.12 & \textbf{88.53} &  &  \small RP   &58.02 & \underline{58.88}& 58.11 & 58.32 &58.86 &57.06  &57.67& 31.58 & \textbf{58.94}\\
 &  \small MAP  &74.94 &76.91& 77.07 & 76.62 &76.51  &\underline{77.09}  &77.56& 69.84 & \textbf{84.49} & &  \small MAP &50.40  & \underline{51.92} & 51.06 & 51.23  &51.38 &50.04 &50.54&22.42 & \textbf{51.93}\\
\hline
\multirow{4}{*}{\rotatebox{90}{ \small \textbf{Avg.}}} & \small  R@1 &94.69  &95.18	& 95.18 & \underline{95.29}   & 94.51&\underline{95.29} &94.79 & 79.97 & \textbf{95.35} & \multirow{4}{*}{\rotatebox{90}{ \small \textbf{Avg.}}} &  \small R@1 &79.79  & 80.57& 80.12 & 80.05 &80.29  &80.37 &\underline{80.58} & 62.49 & \textbf{80.97}\\
 &  \small R@2 &97.57  & \underline{97.96}& 97.65 & 97.81 &97.64 &\textbf{98.01}  &97.60& 89.03 & 97.94 &  & \small  R@2  &85.00 &86.06	 & 85.67 & 85.65 &86.02 &85.51  &\underline{86.21} & 68.96 & \textbf{86.38}\\
 &  \small RP &\underline{70.60} &69.50	& 68.74  & 69.96&69.36 &69.26  &69.29 & 56.88 & \textbf{76.03} &  & \small  RP  &\textbf{41.85} &41.10 & 40.38 & 40.33 &41.15  &39.96  &40.66 &24.15 & \underline{41.76}\\
 & \small  MAP &\underline{61.13 }& 59.87 & 58.75 & 60.49 &59.60 &59.41 &61.50 & 41.90 & \textbf{67.92} & &  \small MAP  &\underline{32.91} & 32.51 & 31.71 &31.65&32.08 &31.40  &32.01 &15.43 & \textbf{33.24}\\

\Xhline{1.5pt}
\end{tabular}    
}    
\label{tab:pacs+}  
\vspace{-3mm}
\end{table*}

\begin{table*}[t]
\centering
\caption{Comparison with the state-of-the-art DML and SDG-DML methods on DomainNet. \colorbox{graytxt}{The left part} corresponds to DML methods, while \colorbox{bluelight}{the right part} belongs to SDG-DML methods. \textbf{Bold} and \underline{underline} indicate the best and second-best results, respectively.
}
\small
\resizebox{\linewidth}{!}{
\setlength{\tabcolsep}{1.6mm}
\begin{tabular}{l|c||cccccccccccccc|cc}
\Xhline{1.5pt}
\multicolumn{2}{c||}{\rotatebox{-90}{{\large \textbf{Method}}}}
& \cellcolor{graytxt}\rotatebox{-90}{\scriptsize\textbf{Marg.~\cite{wu2017sampling}} \scriptsize{[ICCV'17]}}
& \cellcolor{graytxt}\rotatebox{-90}{\scriptsize\textbf{PN~\cite{movshovitz2017no}} \scriptsize{[ICCV'17]}}
& \cellcolor{graytxt}\rotatebox{-90}{\scriptsize\textbf{Cos.~\cite{wang2018cosface}} \scriptsize{[CVPR'18]}}
& \cellcolor{graytxt}\rotatebox{-90}{\scriptsize\textbf{PA~\cite{kim2020proxy}} \scriptsize{[CVPR'20]}}
& \cellcolor{graytxt}\rotatebox{-90}{\scriptsize\textbf{SupC.~\cite{khosla2020supervised}} \scriptsize{[NeurIPS'20]}}
& \cellcolor{graytxt}\rotatebox{-90}{\scriptsize\textbf{Inst.~\cite{zheng2020dual}} \scriptsize{[ACM MM'20]}}
& \cellcolor{graytxt}\rotatebox{-90}{\scriptsize\textbf{DCML~\cite{zheng2021deep}} \scriptsize{[CVPR'21]}}
& \cellcolor{graytxt}\rotatebox{-90}{\scriptsize\textbf{DAS~\cite{liu2022densely}} \scriptsize{[ECCV'22]}}
& \cellcolor{graytxt}\rotatebox{-90}{\scriptsize\textbf{VICReg~\cite{bardes2022vicreg}} \scriptsize{[ICLR'22]}}
&\cellcolor{graytxt} \rotatebox{-90}{\scriptsize\textbf{HSE~\cite{yang2023hse}} \scriptsize{[ICCV'23]}}
& \cellcolor{graytxt}\rotatebox{-90}{\scriptsize\textbf{IDML~\cite{10239539}} \scriptsize{[TPAMI'24]}}
& \cellcolor{graytxt}\rotatebox{-90}{\scriptsize\textbf{DADA~\cite{ren2024towards}} \scriptsize{[AAAI'24]}}
&\cellcolor{graytxt} \rotatebox{-90}{\scriptsize\textbf{PDML~\cite{bhatnagar2025potential}} \scriptsize{[CVPR'25]}}
& \cellcolor{graytxt}\rotatebox{-90}{\scriptsize\textbf{DDML~\cite{park2025deep}} \scriptsize{[AAAI'25]}} 
& \cellcolor{bluelight}\rotatebox{-90}{\scriptsize\textbf{SEE~\cite{yan2023learning}} \scriptsize{[ICCV'23]}}
& \cellcolor{bluelight}\rotatebox{-90}{\scriptsize\textbf{CenterPolar} \scriptsize{[Ours]}} \\
\hline\hline
\multirow{4}{*}{\textbf{R$\to$S}}
 & R@1  & 48.40 & 46.73 & 51.60 & 51.87   & 49.75    &30.64   & 43.04 & 51.44 & 38.51& 50.16 & 52.12 & 52.03 & 50.05 & 52.31&\underline{52.71} & \textbf{53.72}\\
 & R@2  & 56.10 & 54.70 & 59.20 & 59.80  & 57.57    &37.09 & 50.54 & 59.15 &45.64& 58.13 & 59.64 & 59.61 & 57.47 &60.01& \underline{60.38} & \textbf{61.44}\\
 & RP  & 14.03 & 12.98 & 14.45 & 14.32    & \underline{15.46}    &4.09  & 12.30 & 14.85 &7.28& 13.41 & 14.18 & 13.75 & 11.88 &14.82& 15.42 & \textbf{15.84}\\
 & MAP  & 7.11 & 6.35 & 7.72 & 7.28    & 8.21    &1.05  & 6.03 & 8.05 &3.04& 6.78 & 7.54 & 7.31 & 5.95 &7.87 &\underline{8.44} & \textbf{8.94}\\
\hline
\multirow{4}{*}{\textbf{R$\to$I}}
 & R@1& 30.99 & 30.35 & 32.73 & 32.89      & 31.13    &20.65  & 27.04 & 32.36 &25.78& 32.40 & \underline{33.18} & 32.59 & 32.77 &32.72& 32.45 & \textbf{33.60}\\
 & R@2  & 37.64 & 36.92 & 39.11 & \underline{39.61}     & 37.74    &25.81  & 33.12 & 38.43 &31.33& 38.82 & 39.53 & 39.02 & 39.41 &39.45& 38.76 & \textbf{39.68}\\
 & RP  & 7.94 & 7.50 & 7.55 & \underline{7.97}     & \textbf{8.06}    &3.06 & 6.70 & 7.30 &4.34& 7.26 & 7.45 & 6.87 & 6.79 &7.58& 7.47 & 7.70\\
 & MAP & 3.33 & 3.18 & \underline{3.53} & 3.49   & 3.50    &0.66  & 2.76 & 3.34 &1.58 & 3.19 & 3.40 & 3.09 & 2.98 & 3.40&3.49 & \textbf{3.74}\\
\hline
\multirow{4}{*}{\textbf{R$\to$P}}
 & R@1 & 59.73 & 58.89 & 63.57 & 63.30   & 60.93    &40.73  & 54.31 & 62.70&53.98 & 63.20 & \underline{65.85} & 65.62 & 64.83 &64.92& 64.57 & \textbf{67.41}\\
 & R@2& 66.87 & 65.76 & 70.31 & 70.20   & 67.79    &48.07  & 61.01 & 69.51 &60.77& 70.16 & \underline{72.60} & 72.50 & 71.76 & 71.68&71.27 & \textbf{73.87}\\
 & RP  & 21.70 & 21.20 & 23.64 & 23.10    & 22.99    &6.68  & 18.44 & 23.16 &15.64& 22.73 & \underline{25.61} & 25.25 & 22.58 &25.24& 24.74 & \textbf{27.11}\\
 & MAP & 13.43 & 12.78 & 15.29 & 14.49  & 14.68    &1.99  & 10.60 & 15.07 &8.74& 14.25 & \underline{17.00} & 16.86 & 14.38 &16.58 &16.23 & \textbf{18.52}\\
\hline
\multirow{4}{*}{\textbf{R$\to$Q}}
 & R@1 & 34.51 & 33.14 & 38.91 & 39.87   & 36.53   &35.98   & 30.90 & 39.52&33.80 & 36.69 & 38.73 & 38.18 & 35.06 &38.92& \underline{40.02} & \textbf{41.25}\\
 & R@2 & 45.73 & 44.31 & 50.86 & \underline{51.92}   & 47.83    &47.50 & 41.47 & 51.28 &44.88& 48.36 & 50.42 & 49.84 & 46.19 &50.50& 51.73 & \textbf{52.93}\\
 & RP  & 9.58 & 9.31 & 11.37 & 11.96  & 10.61    &10.08  & 8.50 & \underline{12.05} &9.04& 10.11 & 10.97 & 10.56 & 9.10 &11.07 &12.01 & \textbf{12.15}\\
 & MAP  & \textbf{6.00} & 3.19 & 4.44 & 4.66  & 3.97    &3.59  & 2.88 & 4.80 &3.10& 3.63 & 4.17 & 3.94 & 3.14 &4.21& 4.82 & \underline{4.94}\\
\hline
\multirow{4}{*}{\textbf{R$\to$C}}
 & R@1  & 57.31 & 56.34 & 61.38 & 61.38   & 58.66    &36.44  & 50.35 & 61.55 &46.41& 58.94 & \underline{61.96} & 61.27 & 58.00 &61.32& 61.66 & \textbf{63.52}\\
 & R@2 & 65.19 & 64.62 & 69.21 & 69.47     & 66.49   &44.33   & 58.37 & 69.19 &54.21& 67.05 & \underline{69.52} & 68.21 & 66.08 &69.07& 69.25 & \textbf{70.85}\\
 & RP  & 19.70 & 18.89 & 20.71 & 20.38    & 21.25    &6.25  & 17.08 & \underline{21.54}&10.95 & 18.52 & 20.38 & 19.55 & 16.29 &20.57& 21.32 & \textbf{22.51}\\
 & MAP  & 11.59 & 10.88 & 12.69 & 12.02    & 12.91    &2.13 & 9.63 & \underline{13.49} &5.40& 10.85 & 12.61 & 12.00 & 9.45&12.61 & 13.33 & \textbf{14.55}\\
\hline
\multirow{4}{*}{\textbf{Avg.}}
 & R@1  & 46.19 & 45.09 & 49.64 & 49.86    & 47.40   &32.89   & 41.13 & 49.51 &39.70& 48.28 & \underline{50.37} & 49.93 & 48.14 &50.04& 50.28 & \textbf{51.90}\\
 & R@2 & 54.31 & 53.26 & 57.74 & 58.20     & 55.48   &40.56  & 48.90 & 57.51 &47.37& 56.50 & \underline{58.34} & 57.84 & 56.18 &58.14& 58.28 & \textbf{59.75}\\
 & RP & 14.59 & 13.98 & 15.54 & 15.55     & 15.67   &6.03  & 12.60 & 15.78 &9.45& 14.41 & 15.72 & 15.20 & 13.33 &15.86& \underline{16.19} & \textbf{17.06}\\
 & MAP  & 8.29 & 7.28 & 8.73 & 8.39   & 8.65   &1.88   & 6.38 & 8.95 &4.37& 7.74 & 8.94 & 8.64 & 7.18&8.93 & \underline{9.26} & \textbf{10.14}\\
\Xhline{1.5pt}
\end{tabular}
}
\label{tab:DomainNet results}
\end{table*}

\subsubsection{\textbf{Office-Home}}
As shown in the right of Table~\ref{tab:pacs+}, we conducted a systematic evaluation on Office-Home. Our method achieved the best overall performance on Office-Home (mean R@1/R@2/MAP@R: 80.97\% / 86.38\% / 33.24\%), especially MAP, which is 17.81\% higher than SEE (33.24\% vs. 15.43\%). Our method ranked first in the Art domain in R@1, R@2, and MAP@R, with MAP@R also doubling that of the SEE method(29.43\% vs. 13.31\%). Although slightly inferior to the DCML method in the Clipart domain (the gap in MAP is 0.66\%), it is still better than SEE. Similarly, in the product domain, although our R@1 is slightly lower than that of the IDML method (93.25\% vs. 93.88\%), we achieve the best RP and MAP (58.94\% and 51.93\%, respectively). 
Among DML methods, DCML, HIST, and DDML demonstrate strong performance. For example, DCML achieved the best average R@1 (77.79\%) and RP (38.88\%, 28.66\%, and 41.85\%).



Experimental results show that: (1) this method has significant advantages in cross-domain scenarios, and compared with the SEE method, the average MAP@R is improved by more than 60\%; (2) the SEE method performs poorly on datasets with large domain differences (such as PACS and Office-Home), revealing that its agent-based optimization method has generalization limitations; 
(3) on PACS (\textit{e.g.}, in the cartoon domain), our method did not achieve the highest R@1 and R@2 scores. This trade-off arises because CenterPolar, as noted earlier on CUB-200-2011 Ext., focuses on preserving global structural stability rather than pursuing aggressive nearest-neighbor discrimination.

\subsubsection{\textbf{DomainNet}}
Table \ref{tab:DomainNet results} shows that \textbf{CenterPolar} achieved average R@1, R@2, RP, and MAP of 51.90\%, 59.75\%, 17.06\%, and 10.14\%, respectively. These improvements are approximately 1.62\%, 1.47\%, 0.87\%, and 0.88\% compared to the second-place SEE (50.28\%, 58.28\%, 16.19\%, and 9.26\%, respectively). This demonstrates that this method not only improves the accuracy of nearest neighbor retrieval but also significantly enhances the stability of the overall ranking quality (MAP). For domains, \textbf{CenterPolar} achieves the most significant improvements on R$\to$ P: compared to SEE, R@1 improves by 2.84\% (67.41\% vs. 64.57\%) and MAP improves by 2.29\% (18.52\% vs. 16.23\%).  For domains like R $\to$ S and R $\to$ C, which are primarily structure-based and have weak texture information, \textbf{CenterPolar} improves R@1 to 53.72\% and 63.52\%, respectively, representing improvements of 1.01\% (53.72\% vs. 52.71\%) and 1.86\% (63.52\% vs. 61.66\%) over SEE. In contrast, the absolute MAP values for the R $\to$ I and R $\to$ Q domains are smaller (approximately 3.74\% and 4.94\%, respectively). However, \textbf{CenterPolar} still surpasses SEE in R@1 (33.60\% vs. 32.45\%, 41.25\% vs. 40.02\%), respectively. This demonstrates that when the target domain information is presented in a symbolic or extremely simplified form, this method still has some discriminative capability, albeit limited. 
Among DML methods, DAS, IDML, and DDML perform well, with average MAP scores exceeding 8.9\%. IDML shows the best overall performance, ranking second in most metrics. Furthermore, these methods generally outperform earlier proxy-based or contrastive methods, indicating that stronger representation regularization still provides significant benefits in cross-domain transfer scenarios. 
In contrast to traditional DML, SDG-DML achieves better performance, highlighting the effectiveness of its style-generalized modeling ability.

\subsection{Ablation Study}
\label{sub:abl}

\subsubsection{\textbf{Impact of Different Phases}}
Systematic ablation studies (Table~\ref{tab:ablation_cub_cars_transposed}) validate both \textbf{C}$^3$\textbf{E} and \textbf{C}$^4$ phases across three datasets. Individually, each phase improves performance, but with a key trade-off: \textbf{C}$^3$\textbf{E}, only yields higher Recall@1 (25.54\% vs. 24.53\%) and Recall@2 (34.45\% vs. 33.14\%), while the combined phases dominate in RP and MAP@R. This indicates that \textbf{C}$^4$'s intra-class feature compactness slightly reduces local discriminability (affecting Recall), aligning with our objective to optimize overall ranking quality. The \textbf{C}$^4$ induces feature compactness that better preserves semantic relationships across all ranking positions. 
Notably, when neither \textbf{C}$^3$\textbf{E} nor \textbf{C}$^4$ is enabled, the second phase is trained using a standard contrastive loss as the metric objective. The results show that replacing this contrastive objective with \textbf{C}$^4$ leads to consistently better performance. This suggests that, unlike classical prototype-based contrastive learning, \textbf{C}$^4$ jointly consolidates domain-invariant features and refines discriminative metrics: it not only pulls samples toward their class centroids, but also enhances inter-class separability through boundary-aware constraints.

\begin{table*}[t]
\centering
\caption{Ablation study of \textbf{CenterPolar} on CUB-200-2011 Ext., Cars196 Ext., and DomainNet. \textbf{Bold} indicates the best results.}
\resizebox{\linewidth}{!}{
\setlength{\tabcolsep}{1.2mm}
\begin{tabular}{l|l|cccc|cccc||l|l|cccc}
\Xhline{1.5pt}
\multicolumn{2}{c|}{} 
 & \multicolumn{4}{c|}{\textbf{CUB-200-2011 Ext.}} 
 & \multicolumn{4}{c||}{\textbf{Cars196 Ext.}} 
 & \multicolumn{2}{c|}{} 
 & \multicolumn{4}{c}{\textbf{DomainNet}} \\ 
\hline
\multicolumn{2}{c|}{\textbf{C$^3$E}}
 & \XSolidBrush & \XSolidBrush & \CheckmarkBold & \CheckmarkBold
 & \XSolidBrush & \XSolidBrush & \CheckmarkBold & \CheckmarkBold
 & \multicolumn{2}{c|}{\textbf{C$^3$E}}
 & \XSolidBrush & \XSolidBrush & \CheckmarkBold & \CheckmarkBold \\

\multicolumn{2}{c|}{\textbf{C$^4$}}
 & \XSolidBrush & \CheckmarkBold & \XSolidBrush & \CheckmarkBold
 & \XSolidBrush & \CheckmarkBold & \XSolidBrush & \CheckmarkBold
 & \multicolumn{2}{c|}{\textbf{C$^4$}}
 & \XSolidBrush & \CheckmarkBold & \XSolidBrush & \CheckmarkBold \\
\hline
\hline
\multirow{4}{*}{\rotatebox{90}{\small \textbf{R$\to$O}}}
 & \small R@1 & 23.87 & 23.67 & \textbf{25.54} & 24.53 
             & 20.00 & 20.09 & 20.22 & \textbf{20.38}
             & \multirow{4}{*}{\rotatebox{90}{\small \textbf{R$\to$P}}}
             & \small R@1 & 65.49 & 66.60 & 66.83 & \textbf{67.41} \\

 & \small R@2 & 32.73 & 32.63 & \textbf{34.45} & 33.14
             & 28.79 & 29.25 & 28.90 & \textbf{29.43}
             & & \small R@2 & 72.04 & 73.05 & 73.24 & \textbf{73.87} \\

 & \small RP  & 11.38 & 12.06 & 11.71 & \textbf{12.52}
             & 7.16  & 7.32  & 7.30  & \textbf{7.43}
             & & \small RP & 24.28 & 25.90 & 26.12 & \textbf{27.11} \\

 & \small MAP & 4.95 & 5.52 & 5.21 & \textbf{5.79}
             & 1.94 & 2.00 & 2.00 & \textbf{2.07}
             & & \small MAP & 16.26 & 17.66 & 17.88 & \textbf{18.52} \\
\hline
\multirow{4}{*}{\rotatebox{90}{\small \textbf{R$\to$W}}}
 & \small R@1 & 24.14 & 23.73 & 25.02 & \textbf{26.08}
             & 18.53 & 18.90 & 19.19 & \textbf{19.70}
             & \multirow{4}{*}{\rotatebox{90}{\small \textbf{R$\to$C}}}
             & \small R@1 & 63.20 & 63.44 & 63.46 & \textbf{63.52} \\

 & \small R@2 & 34.25 & 33.46 & 35.80 & \textbf{35.96}
             & 27.63 & 27.70 & 27.83 & \textbf{28.51}
             & & \small R@2 & 70.65 & \textbf{70.95} & 70.88 & 70.85 \\

 & \small RP  & 11.40 & 11.74 & 11.96 & \textbf{12.85}
             & 6.22 & 6.30 & 6.29 & \textbf{6.49}
             & & \small RP & 21.93 & 22.04 & 22.12 & \textbf{22.51} \\

 & \small MAP & 4.85 & 5.21 & 5.19 & \textbf{5.84}
             & 1.59 & 1.61 & 1.63 & \textbf{1.69}
             & & \small MAP & 13.99 & 14.12 & 14.17 & \textbf{14.55} \\
\Xhline{1.5pt}
\end{tabular}
}
\label{tab:ablation_cub_cars_transposed}
\end{table*}

\begin{figure*}[t]
  \centering
   \begin{tabular}{cc}
  \includegraphics[width=0.48\linewidth]{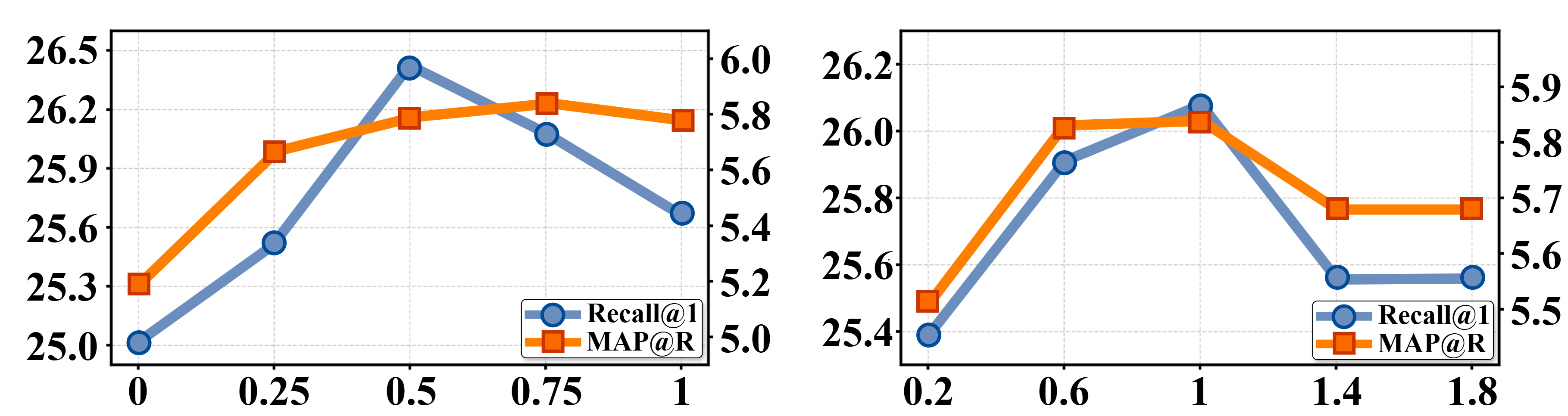} &  \includegraphics[width=0.48\linewidth]{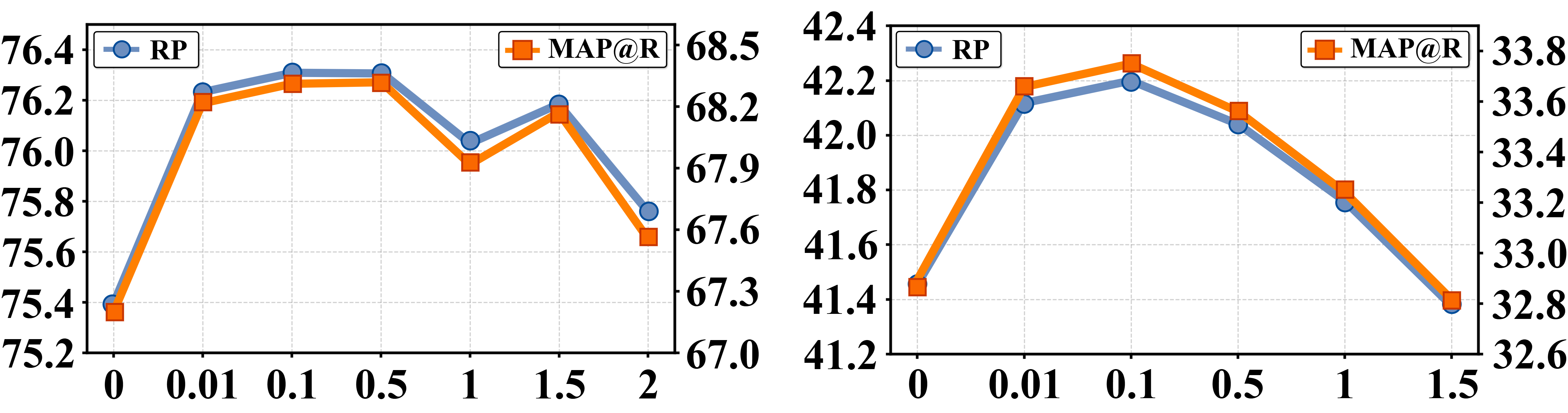}\\
  (a) CUB-200-2011 Ext.  & (b) PACS and Office-Home \\
  \end{tabular}
 \caption{
 Ablation study results of hyperparameters (\(  m \) and \( \lambda \)). The left two parts on  CUB-200-2011 Ext. (R $\rightarrow$ W): Results on different values of \( \lambda \) with \( m = 1 \) and different values of \( m \) with \( \lambda = 0.75 \), respectively. The right two parts show the impact of hyperparameter $\lambda$ when \( m = 1 \) on PACS and Office-Home datasets.
 }
  \label{fig:6}
\end{figure*}

\subsubsection{\textbf{Impact of Hyperparameters}}

\noindent\textbf{(1) CUB-200-2011 Ext.:}
The two parts on the left of Fig.~\ref{fig:6} illustrate the influence of the loss hyperparameters $m$ and $\lambda$ on Recall@1 and MAP@R. In the first graph, we fix $m = 1$ and vary $\lambda \in \{0, 0.25, 0.5, 0.75, 1\}$. The optimal R@1 (26.43\%) occurs at $\lambda = 0.5$, while the peak MAP@R (5.84\%) is achieved at $\lambda = 0.75$. In the second graph, we fix $\lambda = 0.75$ and vary $m \in \{0.2, 0.6, 1, 1.4, 1.8\}$, observing the best overall performance at $m = 1$. As $m$ increases beyond 1, MAP@R steadily declines—dropping by approximately 0.16\%, with a nearly linear downward trend, especially notable between $m = 1$ and $m = 1.4$. This suggests that in controlling the distance between expanded samples and class centroids in the feature space, a moderate margin is essential. Over-enforcing separation (\textit{i.e.}, $m > 1$) may distort the topology of the feature space, weakening intra-class semantic consistency and ultimately degrading the effectiveness of metric learning.

\begin{figure}[t]
  \centering
   \begin{tabular}{cc}
  \includegraphics[width=0.48\linewidth]{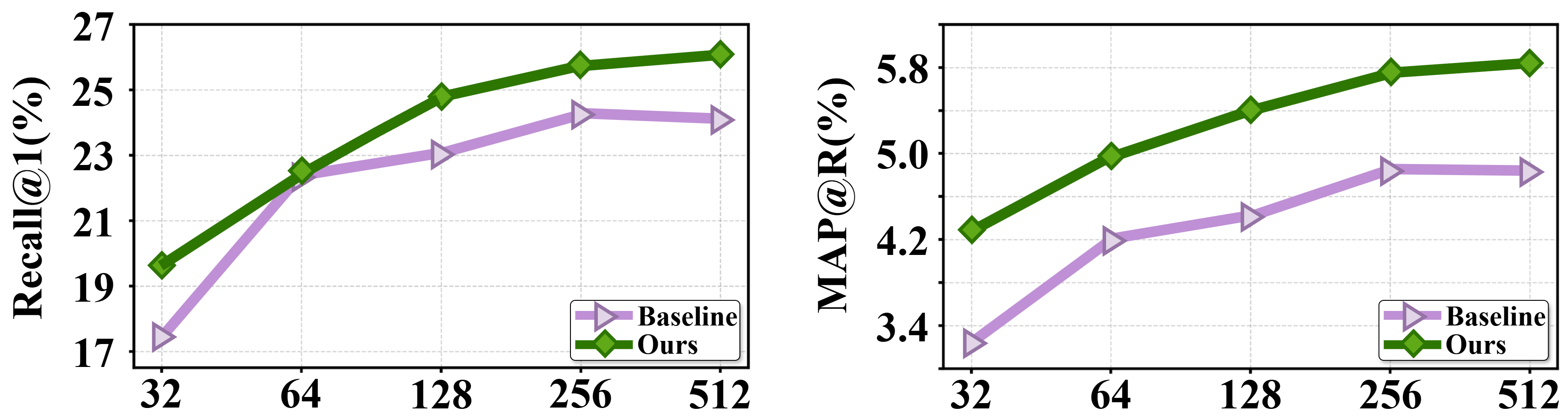}&  \includegraphics[width=0.48\linewidth]{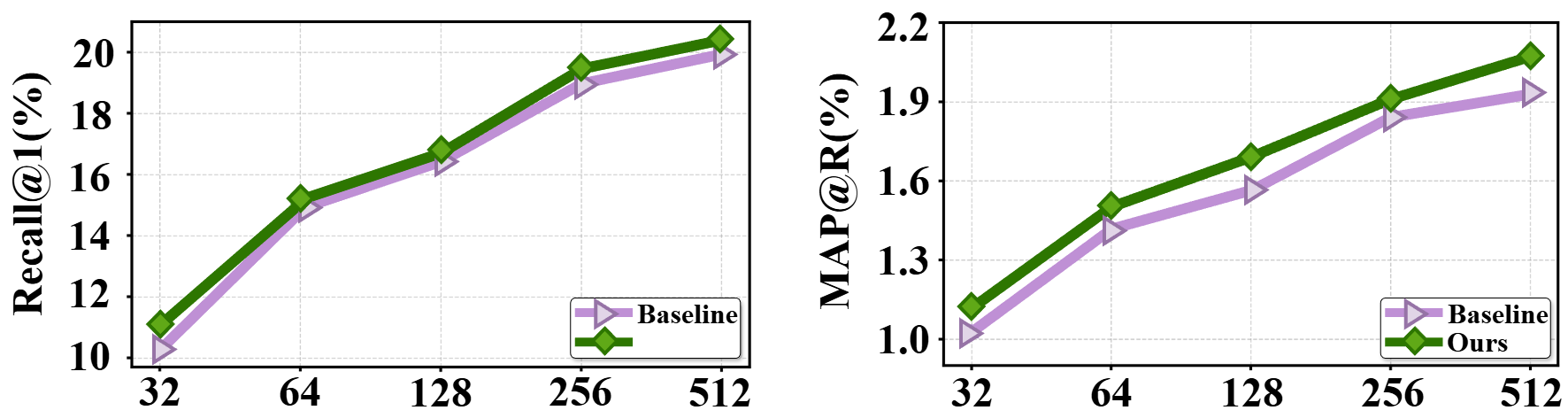}\\
  (a) CUB-200-2011 Ext. & (b) Cars196 Ext.\\
  \end{tabular}
  \vspace{-2mm}
 \caption{
 Ablation study of different embedding dimensions, reporting R@1 and MAP@R. The left two parts present results on CUB-200-2011 Ext. (R $\rightarrow$ W); the right two parts show results on Cars196 Ext. (R $\rightarrow$ O).
 }
  \label{fig:chaocanshu}
  \vspace{-4mm}
\end{figure}

\noindent\textbf{(2) PACS:} The results of different hyperparameters are shown on the left of Fig.~\ref{fig:chaocanshu}. It presents the average performance trend as $\lambda$ increases from $0$ to $2$. The metrics exhibit a concave upward pattern (peaking at $\lambda=0.5$ and then declining). At $\lambda=0$, performance is lowest ($RP=75.39\%$, $MAP@R=67.20\%$). It improves with increasing $\lambda$, peaking at $\lambda=0.5$ ($RP=76.31\%$, $MAP@R=68.32\%$). Beyond $\lambda=0.5$, further increases lead to performance degradation, with minor fluctuations observed at $\lambda=1.5$ ($RP=76.18\%$, $MAP@R=68.16\%$). Crucially, the hyperparameter $m$ was fixed at $1$ for all experiments. Although $m$ is theoretically tunable, empirical results across all five datasets demonstrate that $m = 1$ consistently achieves optimal performance.


\noindent\textbf{(3) Office-Home:} The right side of Fig.~\ref{fig:chaocanshu} shows the results of hyperparameter tuning. We found that as $\lambda$ increases from $0$ to $1.5$, the overall trend is one of first increasing and then decreasing. Both metrics ($RP$ and $MAP$) exhibit a convex curve: performance initially increases and then gradually decreases. Specifically, when $\lambda=0$, the model's performance is lowest, with $RP=41.46\%$ and $MAP=32.88\%$. When $\lambda$ increases to $0.01$, performance improves significantly, rising rapidly. When $\lambda$ increases to $0.1$, both metrics reach their peak, with RP = 42.20\% and $MAP = 33.75\%$. Beyond this point, performance begins to decline, particularly at $\lambda=1.5$, where $RP = 41.38\%$ and $MAP = 32.81\%$. This trend suggests that excessively large $\lambda$ values may cause samples to cluster too closely around the class center, compromising the class discrimination boundary and leading to poor retrieval performance. However, appropriate constraints can enhance the generalization ability.

\subsubsection{\textbf{Impact of Embedding Dimensions}}
Fig.~\ref{fig:6} shows the $R@1$ and $MAP@R$ performance across different embedding dimensions on CUB-200-2011 Ext. and Cars196 Ext. datasets, respectively.
The x-axis represents the embedding dimension, which varies in dimension $\in \{32, 64, 128, 256, 512\}$. Comparing our method with the baseline, we can see that our method outperforms the baseline across all embedding dimensions. Furthermore, our retrieval performance steadily improves with increasing dimensionality, reaching a peak at $512$, with R@1/MAP@R reaching 26.08\%/5.84\% on CUB-200-2011 Ext. and 20.38\%/2.07\% on Cars196 Ext..

\subsubsection{\textbf{Impact of different enhancement methods}}
As shown in Table~\ref{tab:changeC3E}, we replace the \textbf{C$^3$E} phase on CUB-200-2011 Ext. (Water-painting) and Cars196 Ext. (Oil-painting). We added three augmentation methods at different levels: pixel perturbation ColorJitter, feature perturbation Mixup, and adversarial augmentation method StyleGAN2-ADA. ColorJitter modifies the low-level pixel distribution, Mixup blends multiple samples at the feature level to expand the data, and StyleGAN2-ADA generates augmented samples through adversarial training. While each augmentation method improves performance compared to some DML methods, none reaches the performance of our proposed CenterPolar ($R@1 = 26.08\%$ and $MAP = 5.84\%$).

\begin{table}[t]
\centering
\caption{Impact of different enhancement methods. The best results are marked in \textbf{bold}.}
\resizebox{0.8\linewidth}{!}{
\begin{tabular}{l|cccc|cccc}
 \Xhline{1.5pt}
 \multirow{2}{*}{\textbf{Method}} & \multicolumn{4}{c|}{\textbf{CUB-200-2011 Ext.}} & \multicolumn{4}{c}{\textbf{Cars196 Ext.}} \\
 \cline{2-9} 
 & \textbf{R@1} & \textbf{R@2} & \textbf{RP} & \textbf{MAP}& \textbf{R@1} & \textbf{R@2} & \textbf{RP} & \textbf{MAP} \\
\hline
\hline
ColorJitter~\cite{shorten2019survey} + C$^4$ &23.97 & 33.61 &10.96 &4.61 &19.75&28.58&7.17&1.88\\
Mixup~\cite{jin2024survey} + C$^4$ &25.81 &35.80 &12.15 &5.22 &13.69&20.02&4.34&1.05\\
StyleGAN2-ADA~\cite{karras2020training} + C$^4$ &25.66 &35.03 & 12.21&5.48 &20.37&29.38&7.29&2.03\\
\hline
\textbf{CenterPolar} (Ours) &\textbf{26.08} &\textbf{35.96} &\textbf{12.85} &\textbf{5.84} & \textbf{20.38} & \textbf{29.43} & \textbf{7.43} & \textbf{2.07}\\
 \Xhline{1.5pt}
\end{tabular}
}
\label{tab:changeC3E}
\end{table}

\subsection{Visualization} 
\label{sub:vis}
\begin{figure*}[t] 
    \centering
    \includegraphics[width=\linewidth]{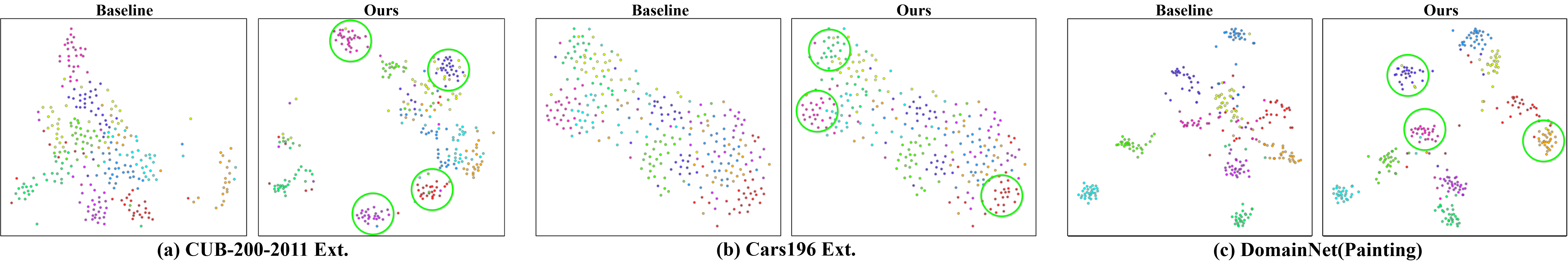}
    \caption{t-SNE visualization of the feature space learned from Baseline and \textbf{CenterPolar} on CUB-200-2011 Ext. (R $\rightarrow$ W),  Cars196 Ext. (R $\rightarrow$ O), and DomainNet (R $\rightarrow$ P) datasets, respectively.
    }
    \label{fig:tsne}
\end{figure*}

\begin{figure}[t]
  \centering
  \includegraphics[width=0.6\linewidth]{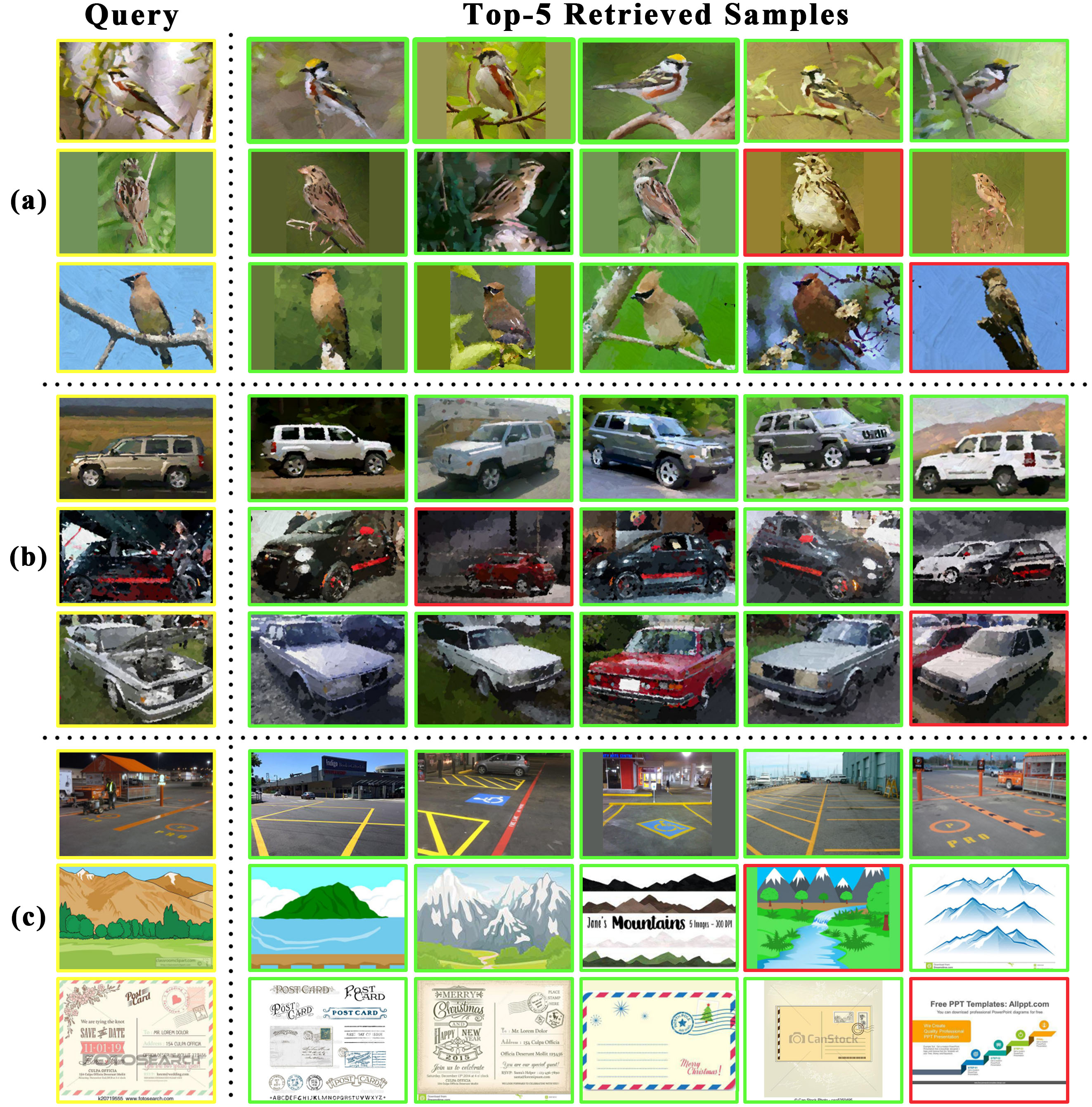}
  \caption{Qualitative retrieval results of the proposed \textbf{CenterPolar} method on (a) CUB-200-2011 Ext., (b) Cars196 Ext., and (c) DomainNet. The true and false matched are marked in green and red boxes, respectively.}
  \vspace{-4mm}
  \label{fig:retrival-results}
\end{figure}

Fig.~\ref{fig:tsne} presents t-SNE visualization results of the learned feature space on the CUB-200-2011 Ext., Cars196 Ext., and DomainNet. For each dataset, we randomly sample 30 instances from 10 categories. Compared to the baseline~\cite{hadsell2006dimensionality}, our method has tighter intra-class compactness and clearer inter-class separation. 
Fig.~\ref{fig:retrival-results} shows the top-5 image retrieval results on three test sets. Experimental results demonstrate that our method accurately returns top-1 matches and maintains relatively stable cross-instance retrieval performance despite complex conditions such as pose variations, lighting differences, color, and perspective shifts. 

\begin{table}[t]
\centering
\caption{Comparison of computational costs between SEE and \textbf{Centerpolar} (Ours).}
\small
\setlength{\tabcolsep}{2.5mm}
\resizebox{0.9\linewidth}{!}{
\begin{tabular}{l|c|cccc}
\Xhline{1.5pt}
\textbf{Dataset} & \textbf{Method} 
& \textbf{Training Time (h)} 
& \textbf{GPU Memory (GB)} 
& \textbf{FLOPs (G)} 
& \textbf{Params (M)} \\
\hline \hline
\multirow{2}{*}{CUB-200-2011 Ext.} 
& SEE                 & 7.5  &20.05   &4.13   & 24.56  \\
& \textbf{Ours} & 2.0  &30.39   &4.13   & 24.56  \\
\hline
\multirow{2}{*}{Cars196 Ext.}
& SEE                 & 11.0 &22   &4.13   & 24.56  \\
& \textbf{Ours} & 1.7 &27.46 &4.13   & 24.56  \\
\Xhline{1.5pt}
\end{tabular}
}
\label{tab:computational cost}
\end{table}

\subsection{Computational Costs}
\label{sub:time}

As shown in Table~\ref{tab:computational cost}, our \textbf{Centerpolar} method achieves significantly faster training than SEE while maintaining comparable FLOPs and parameters. Specifically, training time is reduced from 7.5h to 2.0h on CUB Ext. and from 11.0h to 1.7h on Cars Ext.. The slightly increased GPU memory, due to efficient feature buffering, facilitates faster convergence, further underscoring the practical efficiency of our approach.

\subsection{Limitations}
\label{sub:lim}
While \textbf{CenterPolar} demonstrates good generalization under both domain and category shifts, its ability to address category shift remains limited. We simulate domain shift by performing center expansion via \textbf{C$^3$E}; however, the expanded augmented samples are constrained to seen categories due to the fixed label space. As a result, the model alleviates category shift only indirectly by refining the decision boundaries of seen categories. Future research may explore category-agnostic feature learning or virtual category synthesis. One promising direction is to develop feature representations that decouple category-related and category-invariant components, thereby enhancing generalization. Alternatively, adopting open-set learning paradigms that synthesize virtual categories could enable direct modeling of unseen categories.
Moreover, while SDG-DML explicitly considers domain shift, existing methods, including our \textbf{CenterPolar}, have primarily focused on mitigating style shift, a specific and common form of variation. Notably, domain shift also arises from other factors such as background clutter, illumination changes, and viewpoint variations, which are not explicitly modeled in our current framework. 

\section{Conclusion}
\label{sec:conclusion}

In this paper, we propose \textbf{CenterPolar}, a novel framework for Single-Domain Generalized Deep Metric Learning (SDG-DML) that addresses the dual challenges of category and domain shifts through the class-centric polarization. Our \textbf{CenterPolar} method introduces two collaborative phases: 1) Class-Centric Centrifugal Expansion (\textbf{C$^3$E}) dynamically expands source distributions by pushing samples away from class centroids to enhance domain invariance for unseen domains, and 2) Class-Centric Centripetal Constraint (\textbf{C$^4$}) consolidates discriminative capability by pulling samples toward their class centroids to ensure intra-class compactness and inter-class separation for unseen categories. Extensive experiments on five benchmark datasets demonstrate that \textbf{CenterPolar} achieves superior generalization performance with shorter training time. This work establishes a novel solution for the SDG-DML task, opening avenues for multi-modal extensions. The future work may explore extending this polarization mechanism to multi-modal or federated SDG-DML scenarios. Additionally, the future work will explore more memory-efficient strategies to maintain the fast convergence of CenterPolar while reducing its GPU memory.


\begin{acks}
This research was financially supported by funds from National Natural Science Foundation of China (No. 62376201),
Nature Science Foundation of Hubei Province (No. 2025AFB056),
Key Project of the Science Research Plan of the Hubei Provincial Department of Education (No. D20241103).
Hubei Province Key Laboratory of Intelligent Information Processing and Real-time Industrial System (No. ZNXX2023QNO3), and Fund of Hubei Key Laboratory of Inland Shipping Technology and Innovation (NO. NHHY2023004). 
Key Laboratory of Social Computing and Cognitive Intelligence (Dalian University of Technology), Ministry of Education (No. SCCI2024YB02), 
Research Project of Hubei Provincial Department of Science and Technology (No. 2024CSA075),
and Entrepreneurship Fund for Graduate Students of Wuhan University of Science and Technology (No. JCX2024025). 
\end{acks}

\bibliographystyle{ACM-Reference-Format}
\bibliography{TOMM_references}

\input{work-1/tomm_supplementary}
\end{document}

%% file: work-1/tomm_supplementary.tex
\onecolumn  

\section*{\centering{Supplementary Materials for \\ \emph{Beyond Seen Bounds: Class-Centric Polarization for Single-Domain Generalized Deep Metric Learning\\[30pt]}}}

This is the supplementary material accompanying our main paper, ``Beyond Seen Bounds: Class-Centric Polarization for Single-Domain Generalized Deep Metric Learning''. In this supplementary material, we provide additional information that could not be included in the main paper due to space constraints. We do so in three sections: Section A introduces the datasets; Section B introduces evaluation metrics; Section C compares different tasks; and Section D shows the visualization results. More detailed content is listed as follows:

\begin{itemize}

\item[$\bullet$] A. Datasets 

\item[$\bullet$] B. Evaluation Metrics

\item[$\bullet$] C. Comparison of Different Tasks

\item[$\bullet$] D. Visualization Results
\end{itemize}


\section*{A. Datasets}
\label{sec:Datasets}

\begin{itemize}
    \item 
    \textbf{CUB-200-2011}~\cite{wah2011caltech} comprises 11,788 images spanning 200 bird species. Following the standard Deep Metric learning (DML) protocol, the training set consists of 5,864 images from the first 100 categories, while the testing set consists of the remaining 5,924 images from the last 100 categories. To establish a fine-grained cross-domain benchmark for the SDG-DML task, we adopt the experimental setup of SEE and create an extended dataset variant, denoted as \textbf{CUB-200-2011 Ext.}, utilizing two stylization networks. Specifically, this extended version incorporates two distinct artistic styles: (1) Oil painting style (O) generated by Stylized Neural Painting~\cite{zou2021stylized}, and (2) Watercolor style (W) generated by Paint Transformer~\cite{liu2021paint}. For model training, we employ the original images (R) of the first 100 categories from the CUB-200-2011 dataset as the training set. For evaluation, the test set comprises the stylized images (O and W) corresponding to the last 100 categories from the CUB-200-2011 Ext. dataset. This configuration thereby constructs an evaluation scenario characterized by both category shift (\textit{i.e.}, novel bird species) and domain shift (\textit{i.e.}, novel artistic styles). Fig.~\ref{fig:CUB dataset}(a) shows the dataset details and divisions.
    
    \item 
    \textbf{Cars196}~\cite{krause20133d} includes 16,185 images spanning 196 car categories. Following the standard partitioning protocol, the training set contains 8,054 images from the first 98 categories, while the testing set contains the remaining 8,131 images from the subsequent 98 categories. To fit the SDG-DML setting, we apply the same stylization networks used for \textbf{CUB-200-2011 Ext.} to generate an extended variant of this dataset, denoted as \textbf{Cars196 Ext.}. For model training, we employ the original images (R) of the first 98 categories from the Cars196 dataset. For evaluation, the test set comprises stylized images (O and W) corresponding to the last 98 categories from the Cars196 Ext. dataset. This configuration establishes a consistent evaluation scenario characterized by category shift (\textit{i.e.}, novel car category) and domain shift (\textit{i.e.}, novel artistic styles). Fig.~\ref{fig:CUB dataset}(b) shows the dataset details and divisions.

\begin{figure}[t]
  \centering
  \renewcommand{\thefigure}{A}
  \begin{tabular}{cc}
  \includegraphics[width=0.48\linewidth]{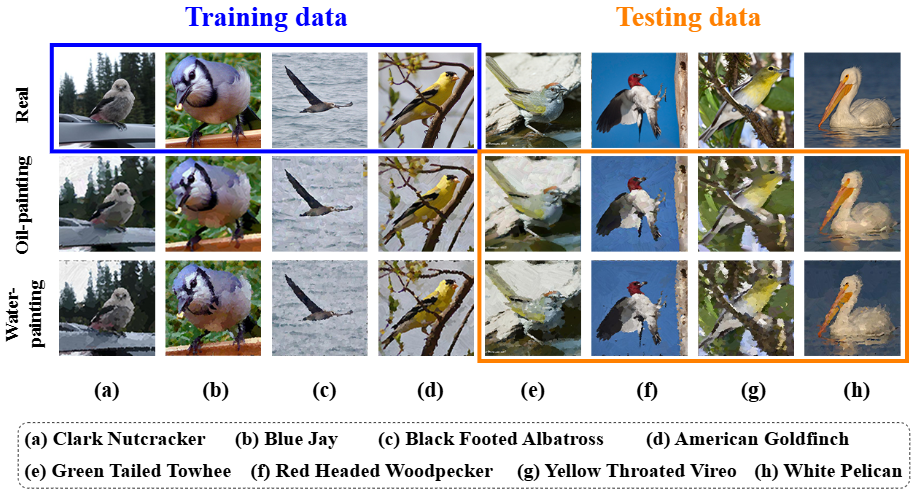} &
  \includegraphics[width=0.48\linewidth]{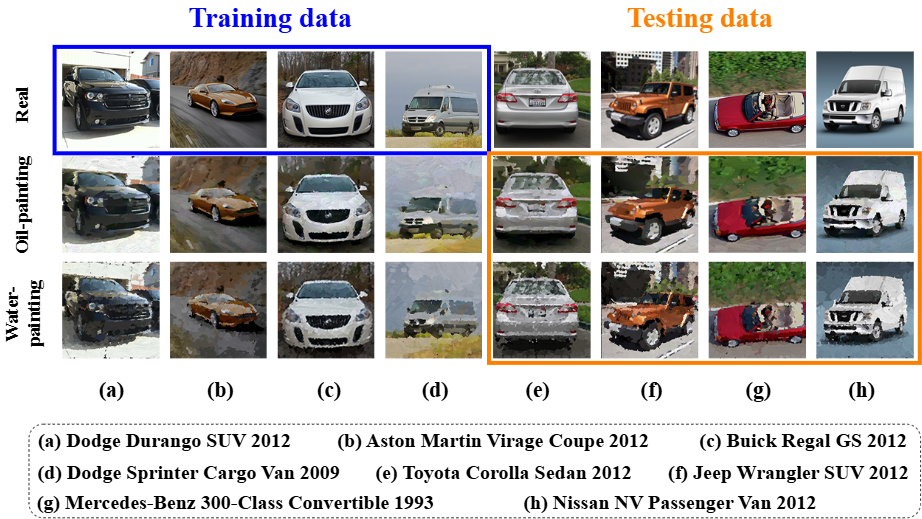}\\
  (a) CUB-200-2011 Ext. &(b) Cars196 Ext.\\
  \includegraphics[width=0.48\linewidth]{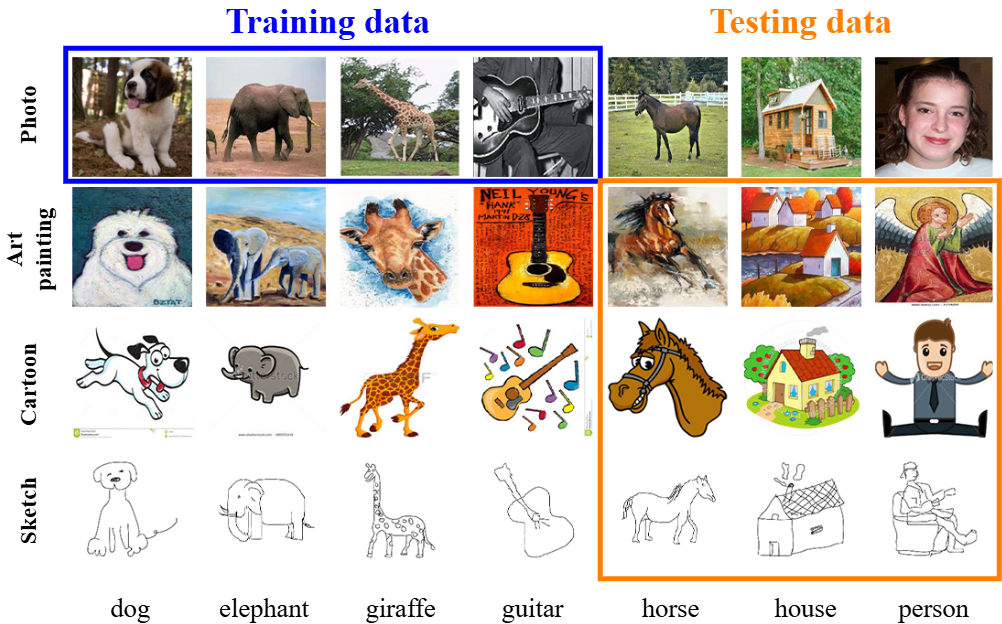} &
  \includegraphics[width=0.48\linewidth]{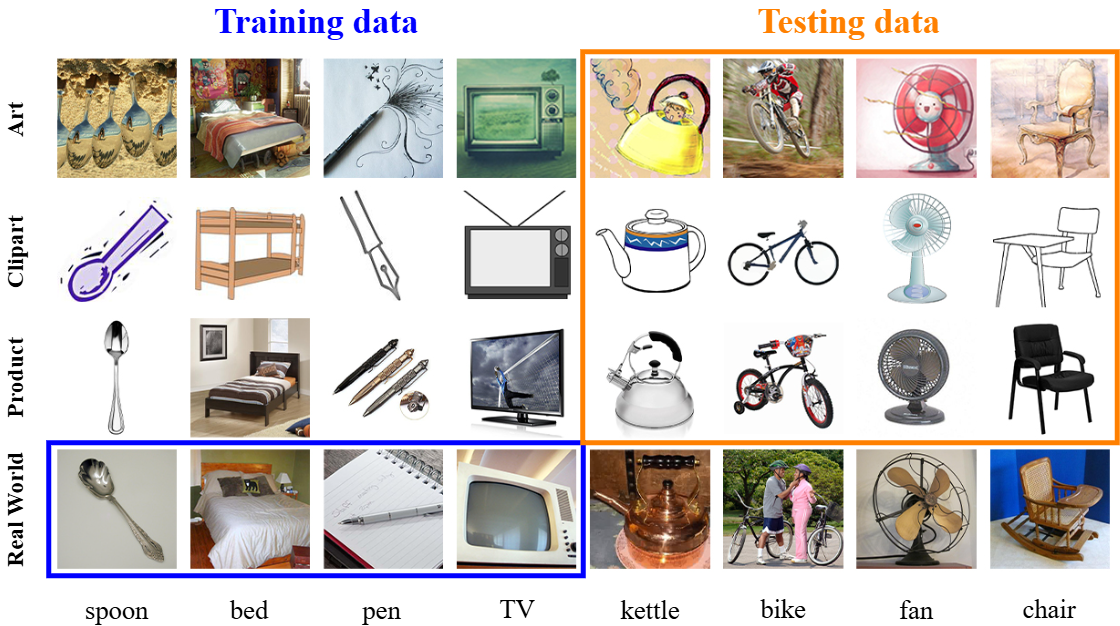} \\
  (c) PACS &(d) Office-Home\\
  \multicolumn{2}{c}{\includegraphics[width=0.48\linewidth]{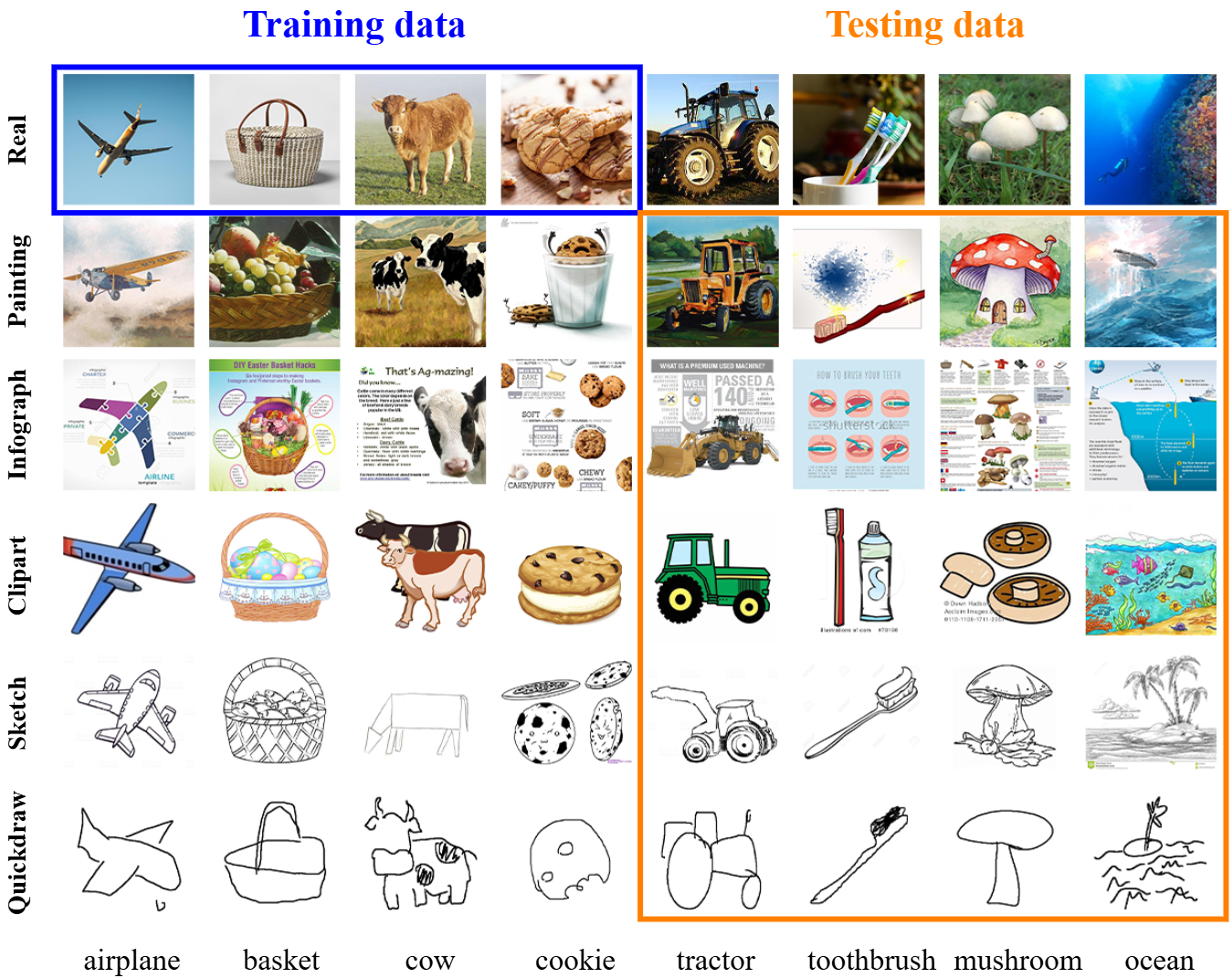}}\\
  \multicolumn{2}{c}{(e) DomainNet}
  \end{tabular}
  \caption{Sample images from CUB-200-2011 Ext., Cars196 Ext., PACS, Office-Home and DomainNet dataset.}
  \label{fig:CUB dataset}
\end{figure}

    \item 
    \textbf{DomainNet}~\cite{peng2019moment} consists of 345 common object categories spanning six distinct domains, with a total of nearly 600,000 images. The six domains include: Clipart (C), Infograph (I), Painting (P), Quickdraw (Q), Real (R), and Sketch (S). Following the standard domain generalization protocol~\cite{oh2016deep}, we employ the Real domain (R) images from the first 173 categories for training. For evaluation, the test set comprises images from the subsequent 172 categories across the remaining five domains: Clipart (C), Infograph (I), Painting (P), Quickdraw (Q), and Sketch (S). This configuration establishes a rigorous evaluation scenario for the SDG-DML task of the main paper under both category shift (\textit{i.e.}, novel object classes) and domain shift (\textit{i.e.}, novel visual styles). Fig.~\ref{fig:CUB dataset}(e) shows the dataset details and divisions.
    
    \item 
    \textbf{PACS}~\cite{li2017deeper} contains 7 object categories across 4 distinct stylistic domains, with a total of nearly 10,000 images. The domains include: Photo (P), Art painting (A), Cartoon (C), and Sketch (S). Following the standard domain generalization protocol, we employ images from the first 4 categories of the Photo domain (P) for training. For evaluation, the test set comprises images from the remaining 3 categories across the other three domains: Art painting (A), Cartoon (C), and Sketch (S). This configuration establishes the SDG-DML benchmark of the main paper characterized by category shift (\textit{i.e.}, novel object classes) and domain shift (\textit{i.e.}, novel artistic styles). Fig.~\ref{fig:CUB dataset}(c) shows the dataset details and divisions.
    
    \item 
    \textbf{Office-Home}~\cite{venkateswara2017deep} comprises 65 object categories spanning 4 diverse domains, with a total of nearly 15,500 images. The domains include: Real World (R), Art (A), Clipart (C), and Product (P). Following the standard domain generalization protocol, we employ images from the first 33 categories of the Real World domain (R) for training. For evaluation, the test set comprises images from the subsequent 32 categories across the remaining three domains: Art (A), Clipart (C), and Product (P). This configuration establishes the SDG-DML benchmark of the main paper characterized by category shift (\textit{i.e.}, novel object classes) and domain shift (\textit{i.e.}, novel visual styles). Fig.~\ref{fig:CUB dataset}(d) shows the dataset details and divisions.
\end{itemize}

\section*{B. Evaluation Metrics}
\label{sub:a.2}
To ensure fair and comprehensive evaluation, we adopt the three standard metrics established in~\cite{musgrave2020metric}: R@$k$, R-Precision (RP), and MAP@R, as follows:

\begin{itemize}
\item  R@$k$ (Recall@$k$) measures the proportion of relevant instances retrieved within the top-$k$ results, widely used in information retrieval and metric learning.

\item  R-Precision (RP) is defined per query as follows:
Let $R$ be the total number of relevant references (same-class instances) in the dataset. When retrieving the top-$R$ nearest neighbors,
\[
    RP = \frac{r}{R}, 
\]
where $r$ denotes the number of relevant instances in this retrieved set~\cite{musgrave2020metric}.

\item MAP@R (Mean Average Precision at R) integrates mean average precision with R-precision:
\[
    \text{MAP@R} = \frac{1}{R} \sum_{i=1}^R P(i) \cdot \mathbb{I}(i), 
\]
where $P(i)$ is the precision at cutoff $i$, and $\mathbb{I}(i)$ is an indicator function equaling 1 if the $i$-th retrieval is relevant (same class), otherwise 0.
\end{itemize}

\section*{C. Comparison of Different Tasks}
\label{sec:motivation}

\begin{figure}[ht]
  \centering
  \renewcommand{\thefigure}{B}
  \includegraphics[width=0.75\linewidth]{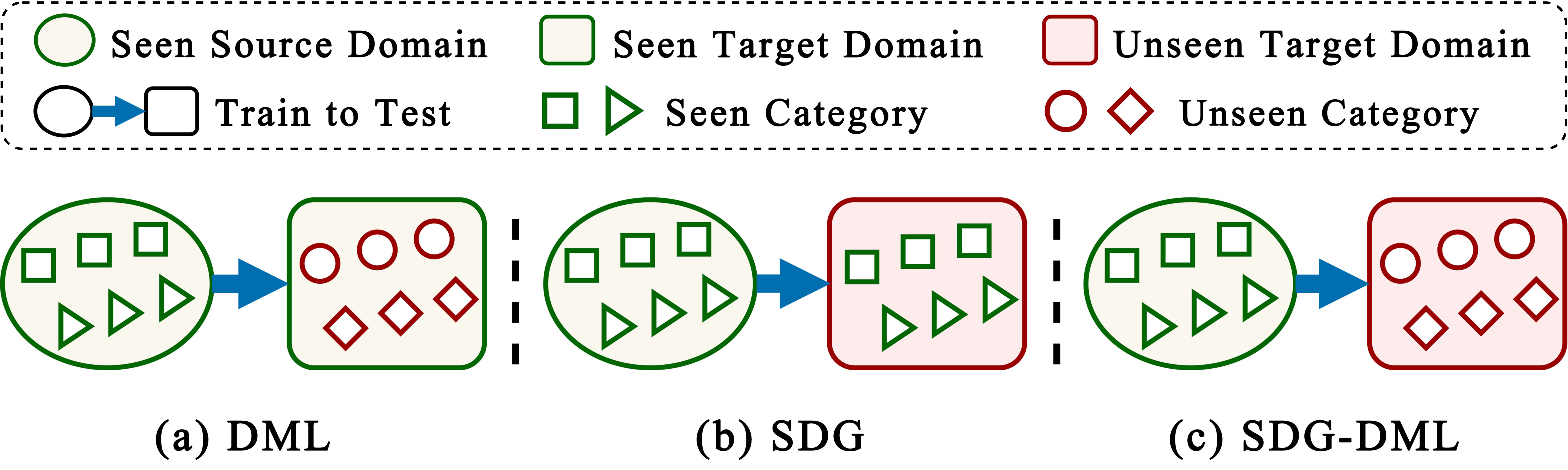}
  \begin{minipage}[c]{\linewidth}
    \centering 
    \vspace{1mm}
    \resizebox{0.7\linewidth}{!}{
    \begin{tabular}{l||cc|cc}
        \Xhline{1.5pt}
        & \multicolumn{2}{c|}{Training Data} & \multicolumn{2}{c}{Testing Data}\\
        \cline{2-5}
        \multirow{-2}{*}{Setting} &  Category & Domain & Category & Domain \\
        \hline
        DML & Seen(\CheckmarkBold) & Seen(\CheckmarkBold) & Unseen(\XSolidBrush) & Seen(\CheckmarkBold) \\
         SDG & Seen(\CheckmarkBold) & Seen(\CheckmarkBold) & Seen(\CheckmarkBold) & Unseen(\XSolidBrush) \\
        SDG-DML & Seen(\CheckmarkBold) & Seen(\CheckmarkBold) & Unseen(\XSolidBrush) & Unseen(\XSolidBrush)\\
        \Xhline{1.5pt}
    \end{tabular}}
\end{minipage}
   \caption{Comparison of different settings in DML, SDG, and SDG-DML. 
   The ellipse and rectangle represent the training data (\textit{i.e.}, source domain) and the testing data (\textit{i.e.}, target domain), respectively. 
   Different shapes in the ellipse and rectangle indicate different classes. 
   In contrast to DML and SDG, SDG-DML learns the knowledge from a single source domain, enabling the metric model to capture domain-invariant patterns that generalize across both unseen categories and domains. (\textit{Best viewed in color}).
   }   
  \label{fig:motivation}
\end{figure}

The critical need for SDG-DML arises from a fundamental representational gap in contemporary metric learning: conventional DML methods achieve strong performance on unseen categories yet remain confined to known domain distributions, while domain adaptation/generalization techniques handle unseen domains but assume fixed category sets. Real-world retrieval systems (such as cross-scenario wildlife monitoring) inherently encounter queries exhibiting simultaneous domain-category novelty, rendering existing approaches deficient: as shown in Figure~\ref{fig:motivation}, DML suffers catastrophic degradation under domain shift (\textit{e.g.}, $>40$\% MAP@R); SDG methods intrinsically lack category-discriminative capability on unseen categories due to closed-set classification objectives. 
SDG-DML bridges this critical gap by establishing the first unified framework that jointly optimizes for:
(i) Domain-invariant metric learning for domain shift.
(ii) Class-discriminative ability for category shift.
Therefore, it enables robust deployment in environments with unpredictable domains and new categories.

\section*{D. Visualization Results}
\begin{figure*}[t] 
  \renewcommand{\thefigure}{C}
    \centering
    \includegraphics[width=0.98\linewidth]{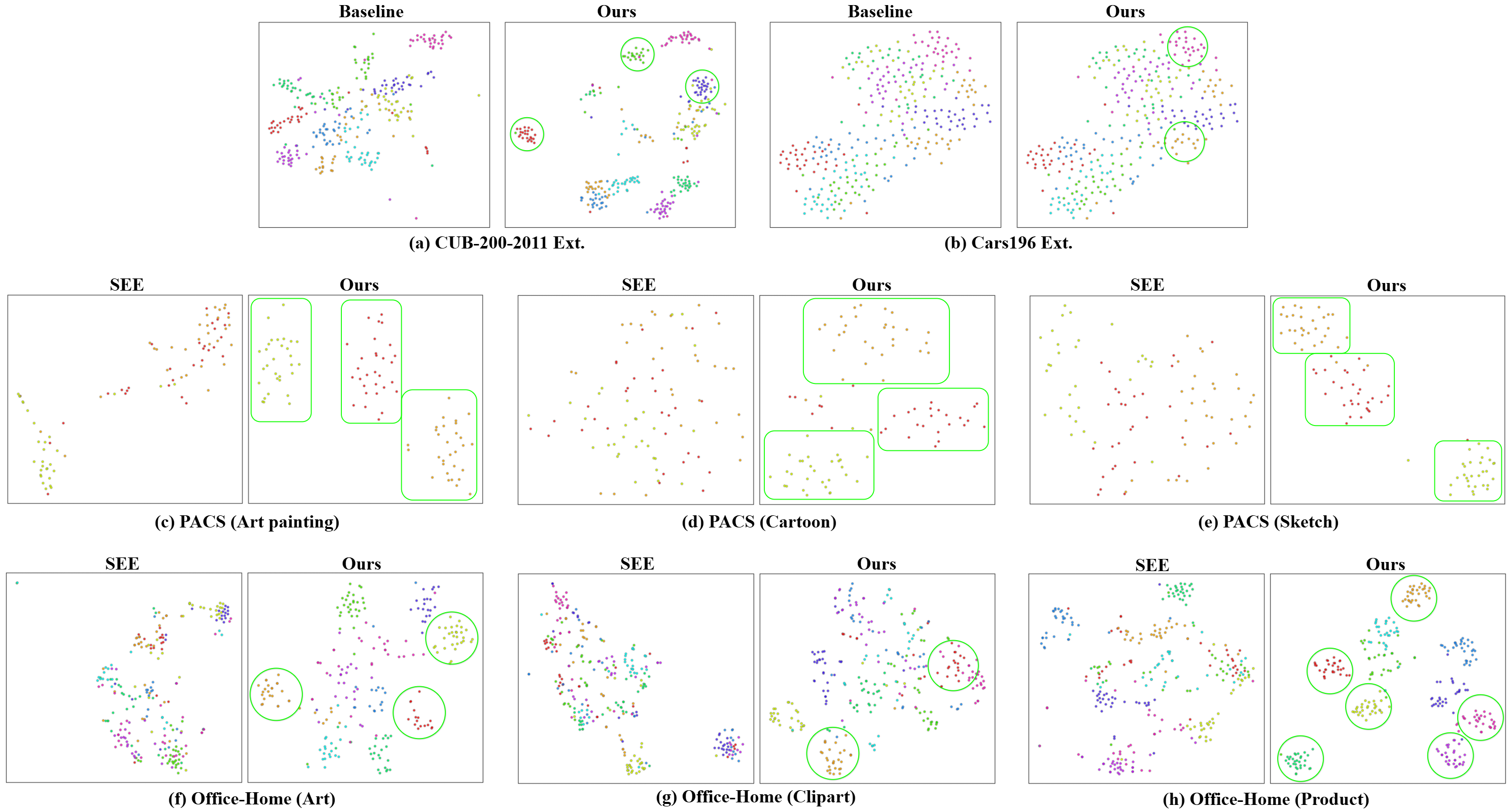}
    \caption{t-SNE visualization of the feature space learned from Baseline and CenterPolar on (a) CUB-200-2011 Ext. (R $\rightarrow$ O), and (b) Cars196 Ext. (R $\rightarrow$ W) datasets, respectively. t-SNE visualization of the feature space learned from SEE and CenterPolar on PACS ((c)-(e)) and Office-Home ((d)-(h)) dataset, respectively.
    }
    \label{fig:tsne++}
\end{figure*}

\begin{figure*}[ht] 
\renewcommand{\thefigure}{D}
    \centering
    \includegraphics[width=0.98\linewidth]{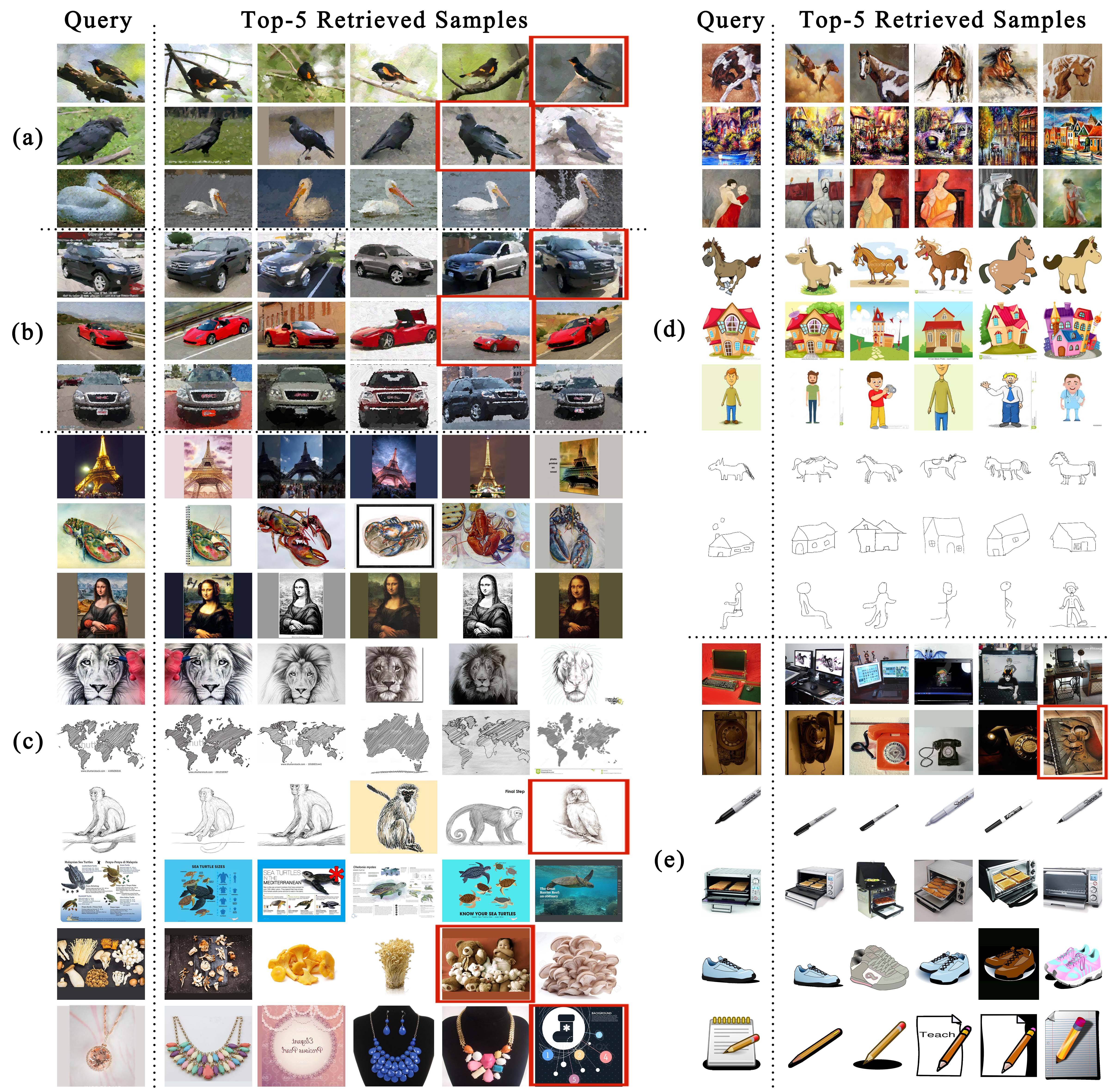}
    \caption{Qualitative retrieval results of the proposed CenterPolar method on (a) CUB-200-2001 Ext., (b) Cars196 Ext., (c) DomainNet, (d) PACS and (e) Office-Home dataset. The false matched images are marked in red boxes, respectively.
    }
    \label{fig:zhanshi2}
\end{figure*}

Fig.~\ref{fig:tsne++} presents t-SNE visualizations of the learned feature space on the CUB-200-2011 Ext., Cars196 Ext., PACS, and Office-Home. For each dataset, we randomly sample 30 instances from 10 categories (the PACS has only 3 categories). Compared to the baseline or SEE, our method has tighter within-class compactness and clearer between-class separation. As illustrated in Fig.~\ref{fig:zhanshi2}, we present the TOP-5 retrieval results across five datasets. Experimental results demonstrate that our method achieves high retrieval accuracy, successfully returning the correct matches for the top-1, top-2, and top-3 positions. Moreover, the method maintains robust cross-instance retrieval performance under challenging conditions, including variations in pose, lighting, color, and perspective.

